\documentclass[lettersize,journal]{IEEEtran}
\usepackage{amsmath,amsfonts}
\usepackage{algorithm}
\usepackage{array}
\usepackage[caption=false,font=normalsize,labelfont=sf,textfont=sf]{subfig}
\usepackage{textcomp}
\usepackage{stfloats}
\usepackage{url}
\usepackage{verbatim}
\usepackage{graphicx}
\usepackage{xcolor}
\usepackage{cite}
\usepackage{algpseudocode}
\usepackage{xcolor}
\usepackage{booktabs}
\usepackage{multirow}

\usepackage{cleveref}
\usepackage{mathrsfs}

\usepackage{mathtools}
\usepackage{amsthm}

\usepackage{multirow}
\theoremstyle{plain}
\usepackage{wrapfig} 
\usepackage{rotating}
\usepackage{pdflscape}

\usepackage{stfloats}
\usepackage{float}
\usepackage{subcaption}
\usepackage{tabularx}
\usepackage{amsmath}
\usepackage{graphicx}
\usepackage{cite}

\begin{document}

\title{\textsc{ChatSR}: Enabling Large Language Models to Understand Scientific Data as They Do Video and Image}

\DeclareRobustCommand*{\IEEEauthorrefmark}[1]{%
    \raisebox{0pt}[0pt][0pt]{\textsuperscript{\footnotesize\ensuremath{#1}}}}
\author{\IEEEauthorblockN{Yanjie Li\IEEEauthorrefmark{1}\IEEEauthorrefmark{2}\IEEEauthorrefmark{3}, 
Lina Yu \textit{Member, IEEE}\IEEEauthorrefmark{1}\IEEEauthorrefmark{5},
Weijun Li \textit{Senior Member, IEEE}\IEEEauthorrefmark{1}\IEEEauthorrefmark{3}\IEEEauthorrefmark{4}\IEEEauthorrefmark{6}*,
Min Wu\IEEEauthorrefmark{1}*, 
Liping Zhang\IEEEauthorrefmark{1},
Jingyi Liu\IEEEauthorrefmark{1}, 
Yusong Deng\IEEEauthorrefmark{1}\IEEEauthorrefmark{4}, 
Mingzhu Wan\IEEEauthorrefmark{1}},
Xin Ning, \textit{Senior Member, IEEE} \IEEEauthorrefmark{1}\IEEEauthorrefmark{4}

\IEEEauthorblockA{\IEEEauthorrefmark{1} AnnLab, Institute of Semiconductors, Chinese Academy of Sciences, Beijing, China}\\
\IEEEauthorblockA{\IEEEauthorrefmark{2} School of Electronic, Electrical and Communication Engineering, University of Chinese Academy of Sciences, Beijing, China}\\
\IEEEauthorblockA{\IEEEauthorrefmark{3} Zhongguancun Academy, Beijing, China}\\
\IEEEauthorblockA{\IEEEauthorrefmark{4} School of Advanced Interdisciplinary Sciences, University of Chinese Academy of Sciences, Beijing 101408, China}\\

\IEEEauthorblockA{\IEEEauthorrefmark{5} College of Materials Science and Opto-Electronic Technology, University of Chinese Academy of Sciences, Beijing, 100049, China}\\

\IEEEauthorblockA{\IEEEauthorrefmark{6} School of Integrated Circuits, University of Chinese Academy of Sciences, Beijing 100049, China}\\

\thanks{*Corresponding author: wjli@semi.ac.cn, yulina@semi.ac.cn,\\dongxiaoli@semi.ac.cn} 
\thanks{This work was supported in part by the National Natural Science Foundation of China under Grant 92370117, in part by CAS Project for Young Scientists in Basic Research under Grant YSBR-090.}}
\markboth{Journal of \LaTeX\ Class Files,~Vol.~14, No.~8, August~2021}%
{Shell \MakeLowercase{\textit{et al.}}: A Sample Article Using IEEEtran.cls for IEEE Journals}

\maketitle

\begin{abstract}

Current multimodal large language models (MLLMs) are mainly focused on the understanding and processing of perceptual modalities such as images and videos, while their capability for scientific data understanding remains insufficient. To this end, we propose ChatSR, a novel multimodal large language model tailored for scientific data understanding. ChatSR treats scientific data as a new modality analogous to visual content and, through carefully designed encoders and modality alignment mechanisms, maps scientific data into a representation space that can be processed by large language models, enabling the model to grasp the structural characteristics and underlying regularities of scientific data. Building on this foundation, ChatSR further exploits the rich domain knowledge and strong reasoning abilities of large language models to emulate a knowledgeable human scientist: based on user-specified prior constraints and preferences expressed (such as requirements on periodicity, symmetry, etc.), it automatically generates mathematical formulas that not only accurately fit the observed data but also conform to domain priors, thereby characterizing the latent laws embodied in scientific data and promoting the automation of scientific discovery.

Experiments on 13 datasets show that ChatSR achieves state-of-the-art performance on traditional symbolic regression benchmarks. In addition, ChatSR exhibits a promising zero-shot ability to understand and utilize types of prior knowledge that are not present in its training data.
\end{abstract}

\begin{IEEEkeywords}
Multi-modal Large Language Model, Symbolic Regression, Scientific discovery, Mathematical modeling.
\end{IEEEkeywords}

\section{Introduction}
\IEEEPARstart{M}{athematical} formulas are the language through which humans communicate with nature. Owing to their concise form, they can capture the potential relationships among the variables involved.  A central goal of scientific research is to derive compact expressions from observational data that reveal the physical laws underlying observed phenomena.  However, manually discovering such formulae typically requires a long period of time and substantial expertise. This has motivated efforts to use artificial intelligence methods to enable computers to automatically discover mathematical formulas from data, a task commonly known as symbolic regression. This is where the symbolic regression problem comes in. 

Specifically, given observations $D = \{X, y\}$, symbolic regression (SR) seeks a function $f$ such that $y \approx f(X)$, where $X \in \mathbb{R}^{n \times d}$ is the input matrix, $y \in \mathbb{R}^n$ is the target vector, $d$ is the dimensionality of the input, and $n$ is the number of data points. The function $f$ is constructed from a set of primitive operators and operands such as $+,-,\times,\div,\sin,\cos$, constants $C$, and variables $x_1,\dots$. An expression is typically represented as a binary tree and then linearized by a preorder traversal to obtain a sequence of symbols. For example, the expression $y = 4.2 \sin(x_1)$ can be represented as the symbol sequence $[\times, C, \sin, x_1]$.

Traditional methods treat symbolic regression as a combinatorial optimization problem and address it using algorithms such as genetic programming (GP) and reinforcement learning.  These methods are typically trained from scratch on each new dataset, which makes them robust to noise and broadly applicable.

However, in real scientific practice, researchers often possess substantial prior knowledge or assumptions when modeling observational data with analytic expressions.  For example, suppose we use time $t$ as the variable and seek a function $f(t)$ that models the intensity of light over the course of a month.  We know a priori that $f(t)$ should be approximately periodic, so we would expect $f(t)$ to be periodic or at least to involve sinusoidal components such as $\sin(\cdot)$.

In recent years, multimodal large language models (MLLMs), exemplified by LLaVA~\cite{llava} and Qwen2.5-VL~\cite{qwen2.5-vl}, have made remarkable progress. MLLMs are able to answer user queries based on inputs from various modalities such as images, PDF files, and videos. For example, given an image containing a person and a dog and the query ``What is in the image?'', an MLLM may respond, ``There is a person and a dog.'' If we refine the query to ``What pets are in the image?'', the model may answer, ``There is a pet dog in the image.''

This naturally raises the question of whether we can develop a multimodal large language model specifically for symbolic regression. In such a setting, we could provide observational data $(X, y)$ together with a prompt such as ``Please generate an expression that fits this data,'' and the model would output a corresponding analytic expression. Furthermore, if we include prior knowledge in the prompt, for example, ``Please generate a \textbf{periodic} expression that fits this data,'' the model should produce an expression that is periodic.

In this paper, we propose \textsc{ChatSR}, a multimodal large language model for symbolic regression that inherits the strong knowledge and language understanding capabilities of modern large language models. Given observational data $(X, y)$ and a natural-language description of prior knowledge or modeling requirements, \textsc{ChatSR} can automatically generate analytic expressions that both fit the data and satisfy the specified constraints.

We summarize our contributions as follows:
\begin{itemize}
\item We present \textsc{ChatSR}, a multimodal large language model tailored to symbolic regression. After providing scientific data as input, users can describe arbitrary prior knowledge and assumptions in natural language, and \textsc{ChatSR} generates expressions that conform to these requirements.

\item We propose and construct the optimal chain of inference(OCOI), which gives an expressive inference chain with increasing $R^2$, from which we construct multi-turn dialogue data for symbolic regression multimodal large language models.
\item We introduce a new benchmark, \textsc{Knowledge}, which contains 50 expressions exhibiting a variety of common properties. It can be used to evaluate the ability of symbolic regression methods to generate expressions consistent with given prior knowledge.

\item We point to new research directions for leveraging multimodal large language models in symbolic regression and scientific discovery.

\end{itemize}

\section{Relation work}

\subsection{Multi-modal Large Language Models}

Recently, models such as CLIP \cite{CLIP} and ALIGN \cite{ALIGN} have been pre-trained on noisy image–text pairs from the web using contrastive loss, which is recognized as one of the most effective methods for feature learning \cite{he2020momentum,chen2020simple,li2020prototypical,li2020mopro}. These models achieve remarkable performance on image–text retrieval tasks but are limited in their ability to model more complex interactions between images and text necessary for other vision-and-language (V+L) tasks \cite{kim2021vilt}, such as visual question answering (VQA) \cite{antol2015vqa}. Subsequent studies \cite{wang2021simvlm,wang2022ofa,piergiovanni2022answer} have introduced encoder–decoder frameworks trained with generative objectives, demonstrating robust performance across diverse V+L benchmarks; simultaneously, their visual encoders maintain competitive accuracy on image classification. A parallel line of research \cite{singh2022flava,li2021align,li2022blip,chen2023pali,liu2024visual} explores unifying image and text representations, typically via multi-stage pretraining of unimodal and multimodal modules. For instance, ALBEF \cite{li2021align} employs a dual-encoder architecture that couples contrastive learning with masked language modeling (MLM) to enhance efficiency. CoCa \cite{yu2022coca} trains an image–text foundation model from scratch in a single pretraining stage, unifying contrastive and generative objectives. BEITv3 \cite{beit3} treats images as a form of language via a mapping layer before integrating them with encoded text into a large GPT-style model. LLaVA \cite{llava} aligns image features to the token space of a large language model (LLM) \cite{llm1,llm2,llm3,llm4,llm5}, concatenating visual and textual tokens to enable instruction following in vision–language settings.
Instruction-tuned MLLMs such as InstructBLIP extend BLIP-2 with multimodal instruction corpora and an instruction-aware query transformer, yielding strong zero-shot transfer across V+L tasks \cite{dai2023instructblip}. To better handle interleaved inputs, LLaVA-NeXT-Interleave unifies multi-image and video understanding within a single training recipe, improving temporal and spatial reasoning while maintaining image QA performance \cite{li2024llava}. Concurrently, Google’s PaliGemma and PaliGemma~2 streamline open V+L pretraining via a SigLIP-style vision tower and lightweight decoders, providing competitive open baselines for captioning and VQA under efficient training regimes \cite{paligemma1,paligemma2}. The Qwen2-VL introduces dynamic-resolution visual tokenization and M-RoPE for joint text–image/video positional fusion, scaling from 2B–72B parameters and narrowing the gap to proprietary systems on diverse multimodal suites \cite{qwen2-vl}. Building on this, Qwen2.5-VL reports gains in document parsing (layout/structure), precise localization, and long-video comprehension with native dynamic-resolution ViT and windowed attention, alongside improved agentic tool use \cite{qwen25-vl}. Most recently, Qwen3(Qwen3-VL) generalizes the series with larger ``Instruct''/``Thinking'' variants and broader capability coverage—multi-image and video understanding, advanced OCR and grounding, long-document parsing, and extended-context multimodality—delivering competitive results versus state-of-the-art closed models and offering practical deployment checkpoints \cite{qwen3}.

\subsection{Symbolic Regression}
\paragraph{Symbolic Regression Based on Genetic Programming}
This kind of method is a classical kind of algorithm in the field of symbolic regression. GP \cite{10.1145/2576768.2598291}, \cite{McConaghy2011}, \cite{7473913} is the main representative of this kind of method; its main idea is to simulate the process of human evolution. Firstly, it initialized an expression population, then generated new individuals by crossover and mutation, and finally generated a new population by fitness. The above process is repeated until the target expression is obtained. RSRM\cite{xureinforcement} integrates the GP algorithm with Double Q-learning\cite{hasselt2010double} and the MCTS algorithm\cite{coulom2006efficient}. A Double Q-learning block, designed for exploitation,  that helps reduce the feasible search space of MCTS via properly understanding the distribution of reward. In short,  the RSRM model consists of a three-step symbolic learning process: RL-based expression search, GP tuning, and MSDB. In this paper\cite{fong2022rethinking}, the fitness function of the traditional GP algorithm is improved, which promotes the use of an adaptability framework in evolutionary SR that uses fitness functions that alternate across generations. LLM-SR\cite{llmsr1} and ICSR\cite{llmsr2} use the large language model to aid the search process, just let the LLM produce a series of expressions, and then use the good expressions as a hint to let the LLM continue to produce a new batch of expressions until the target expression is reached.
\begin{figure*}[ht]
\centering
\hspace*{-0.45cm}
\includegraphics[width=170mm]{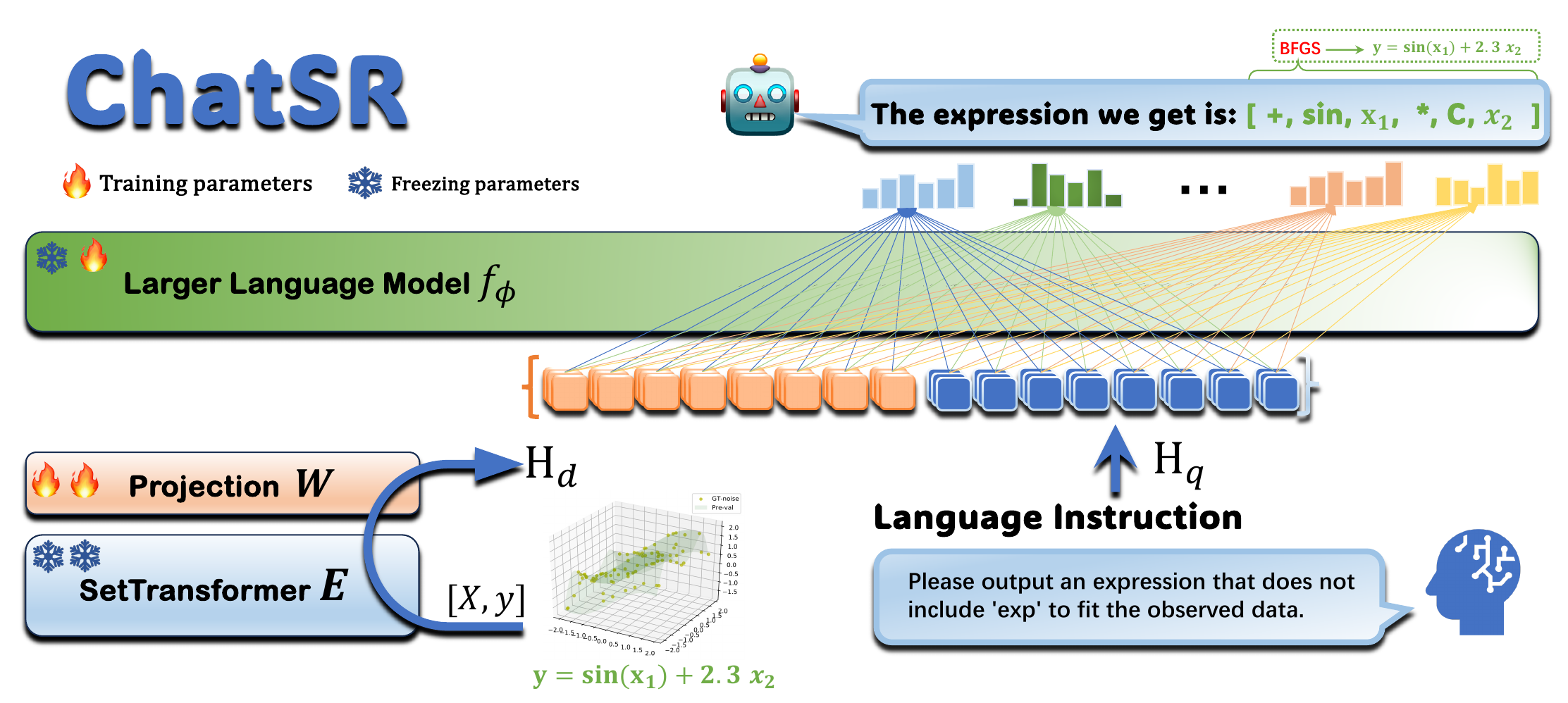}
\caption{
This figure shows a schematic diagram of the overall process of ChatSR.
}
\label{fig1}
\end{figure*}
\paragraph{Symbolic Regression Based on reinforcement learning}
Reinforcement learning-based algorithms treat symbolic regression as a combinatorial optimization problem. The typical algorithm is DSR\cite{petersen2019deep}, which uses a recurrent neural network as a policy network to generate a probability distribution P for sampling, and then samples according to the probability P to obtain multiple expressions. The reward value of the sampled expressions is calculated, and the policy network is updated with the risky policy and the loop continues until the target expression is obtained. DSO\cite{mundhenk2021symbolic} is based on DSR by introducing the GP algorithm. The purpose of the policy network is to generate a better initial population for the GP algorithm. Then, the risk policy gradient algorithm is also used to update the policy network. Although the above two algorithms are very good, the efficiency is low, and the expression is more complex, especially the DSO algorithm is more obvious. There have been many recent symbolic regression algorithms based on the Monte Carlo tree search. SPL\cite{sun2022symbolic} uses MCTS in the field of symbolic regression and introduces the concept of modularity to improve search efficiency. However, due to the lack of guidance from MCTS, the search efficiency of this algorithm is low. To improve the search efficiency of the algorithm, the two algorithms DGSR-MCTS \cite{kamienny2023deep} and TPSR \cite{shojaee2024transformer} introduced the policy network to guide the MCTS process based on the previous algorithm. While maintaining the performance of the algorithm, it greatly improves the search efficiency of the algorithm. However, although the above two algorithms improve the search efficiency of the algorithm, they reduce the Versatility of the algorithm, and the noise robustness ability of the algorithm is also greatly reduced. To solve the above problems and balance the Versatility and efficiency of the algorithm, SR-GPT \cite{li2024discovering} uses a policy network that learns in real-time to guide the MCTS process. It achieves high performance while being efficient in search.

\paragraph{Symbolic Regression Based on Pre-training}
Many SR methods based on reinforcement learning have good Versatility. However, its search efficiency is relatively low, and it often takes a long time to get a good expression. In contrast, pre-trained models treat the SR problem as a translation problem and train a transformer with a large amount of artificially synthesized data in advance. Each prediction only needs one forward propagation to get the result, which is relatively efficient. SymbolicGPT\cite{valipour2021symbolicgpt} was the first large-scale pre-trained model to treat each letter in a sequence of symbols as a token (e.g.['s',' i','n', '(', 'x', ')']). A data feature extractor is used as the encoder, and then each token is generated by the Decoder in turn. Finally, the predicted sequence and the real sequence are used for cross-entropy loss. BFGS is used to optimize the constant at placeholder 'C'. NeSymReS\cite{biggio2021neural} builds on symbolicGPT by not thinking of each individual letter in the sequence of expressions as a token. Instead, Nesymres represents the expression in the form of a binary tree, which is then expanded by preorder traversal, and considers each operator as a token (e.g., ['sin','x']). Then, SetTransformer is used as the Encoder of the data, and finally, Decoder is used to generate the expression sequence. The overall framework and idea of the End-to-End \cite{kamienny2022end} algorithm are not much different from NeSymReS, but End-to-End abandons the constant placeholder 'C', encodes the constant, and directly generates the constant from the decoder. The constants are then further optimized by Broyden-Fletcher-Goldfarb-Shanno (BFGS) \cite{liu1989limited}. Based on End-to-End, NSRwH\cite{bendinelli2023controllable} tries to apply some prefixes to prompt the model to generate expressions that conform to the prior. But the effect is not obvious.
Symformer\cite{vastl2024symformer} is slightly different from the previous pre-trained models in that it directly generates the constant values in the expression as well as the sequence of expressions. LLM-SR\cite{llmsr1} and ICSR\cite{llmsr2} use LLM as a guide, but do not train LLM specifically for SR tasks. SNIP\cite{meidani2023snip} first applies contrastive learning to train the feature encoder and then freezes the encoder to train the decoder. But SNIP works well only when combined with a latent space optimization
(LSO)\cite{bojanowski2017optimizing} algorithm.
\begin{figure*}[htp]
\centering
\hspace*{0.11cm}
\includegraphics[width=166mm]{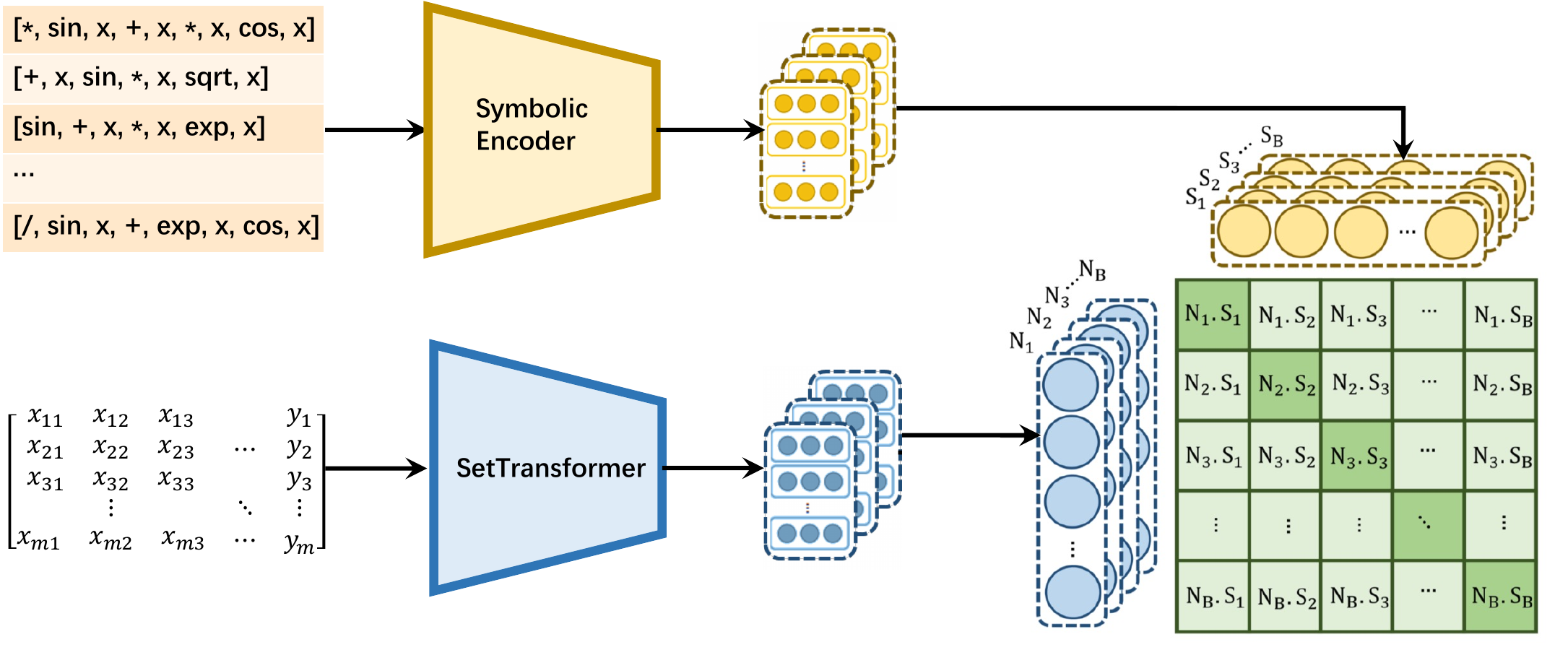}
\caption{
The feature extractor Setransformer is pre-trained by contrastive learning.
}
\label{contransive}
\end{figure*}
MMSR\cite{li2024mmsr} solves the symbolic regression problem as a pure multimodal problem, takes the input data and the expression sequence as two modalities, introduces contrastive learning in the training process, and adopts a one-step training strategy to train contrastive learning with other losses.

\paragraph{Symbolic Regression Based on Deep Learning}
This class of methods combines symbolic regression problems with artificial neural networks, where EQL replaces the activation function in ordinary neural networks with [sin, cos,...] and then applies pruning methods to remove redundant connections and extract an expression from the network. EQL\cite{kim2020integration} is very powerful; however, it can't introduce division operations, which can lead to vanishing or exploding gradients.
The main idea of the AI Feynman 1.0 \cite{udrescu2020ai} and AI Feynman 2.0\cite{udrescu2020ai} series algorithms is to “Break down the complex into the simple” by first fitting the data with a neural network, and then using the trained neural network to discover some properties (e.g., Symmetry, translation invariance, etc.) to decompose the function hierarchically. AI Feynman 2.0 introduces more properties based on AI Feynman 1.0, which makes the scope of its application more extensive relative to AI Feynman 1.0. MetaSymNet\cite{li2023metasymnet} takes advantage of the differences between symbolic regression and traditional combinatorial optimization problems and uses more efficient numerical optimization to solve symbolic regression.

\section{Method}
We generated a total of 5M expressions, based on which we generated 30M Q$\&$A training data about expressions. Each piece of data contains observation data, [X, Y], and a text question-answer pair. 


We generate a corpus of 5M expressions and, based on this corpus, construct 30M question--answer (Q\&A) training instances about expressions. Each instance consists of observational data $(X, Y)$ and one or more rounds of textual question--answer.

We first train a SetTransformer as the data feature encoder $E$ of \textsc{ChatSR} using contrastive learning on 5M pairs of $(X, Y)$ and their corresponding expression preorder traversals (e.g., $[\sin, \times, x, x]$)~\cite{meidani2023snip}. The training procedure is illustrated in Fig.~\ref{contransive}. We then freeze the parameters of both the SetTransformer and the LLM, and pre-train only the projection layer that maps data features into the text feature space. Finally, we freeze only the SetTransformer and jointly fine-tune the projection layer and the LLM. The overall architecture of \textsc{ChatSR} is shown in Fig.~\ref{fig1}.

\subsection{Expressions generation}
\label{section3.1}

In \textsc{ChatSR}, we adopt the following set of primitive symbols:
\[
\{+, -, \times, \div, \sin, \cos, \log, \sqrt{\cdot}, C, x_1, x_2, \dots, x_n\}.
\]
Here, $C$ denotes a placeholder for a numeric constant, and $x_1, \dots, x_n$ denote variables. Expressions composed of these symbols can be represented as binary expression trees. By performing a preorder traversal (root--left--right) of such a tree, we obtain a sequence of symbols. For example, the expression $\sin(2.6x)$ can be written as $\sin(C \times x)$, whose preorder traversal is $[\sin, \times, C, x]$. 

To synthesize training data, we first generate a preorder sequence of an expression by randomly sampling symbols from the primitive set. We then decode this sequence back into an expression, and evaluate it on input $X$ to obtain the corresponding observations $(X, y)$.

\subsubsection{Generation stopping criterion: $count = 0$}
\label{stop_decision}
To determine when expression generation should stop, we introduce a counter variable $count$, initialized to $1$. In addition, we define an $\text{Arity}(s)$ function: if $s$ is a binary operator (e.g., $+, -, \times, \div$), then $\text{Arity}(s) = 2$; if $s$ is a unary operator (e.g., $\sin, \cos, \dots$), then $\text{Arity}(s) = 1$; and if $s$ is a variable ($x_1, \dots, x_n$) or the constant placeholder $C$, then $\text{Arity}(s) = 0$.

We iteratively generate symbols in preorder. At each step, we randomly select a symbol $s$ from the symbol set and update the counter according to
\[
count \leftarrow count - 1 + \text{Arity}(s).
\]
This process is repeated until $count = 0$, at which point we obtain a complete preorder traversal of an expression tree.

\subsubsection{Generation constraints}
To ensure that the generated expressions are meaningful, we impose the following constraints:
\begin{itemize}
\item[(1)] Trigonometric functions are not allowed to be nested (e.g., $\sin(\cos(x))$), as such forms are rarely encountered in practical scenarios.
\item[(2)] For functions such as $\log(x)$ and $\sqrt{x}$, the argument $x$ must not be negative. For example, expressions like $\log(\sin(x))$ and $\sqrt{\cos(x)}$ are considered invalid, since $\sin(x)$ and $\cos(x)$ can take negative values.
\end{itemize}

\subsection{Training data collection}
\subsubsection{Single-turn dialogues}
\begin{figure*}[htp]
\centering
\vspace*{-0.9cm}
\includegraphics[width=184mm]{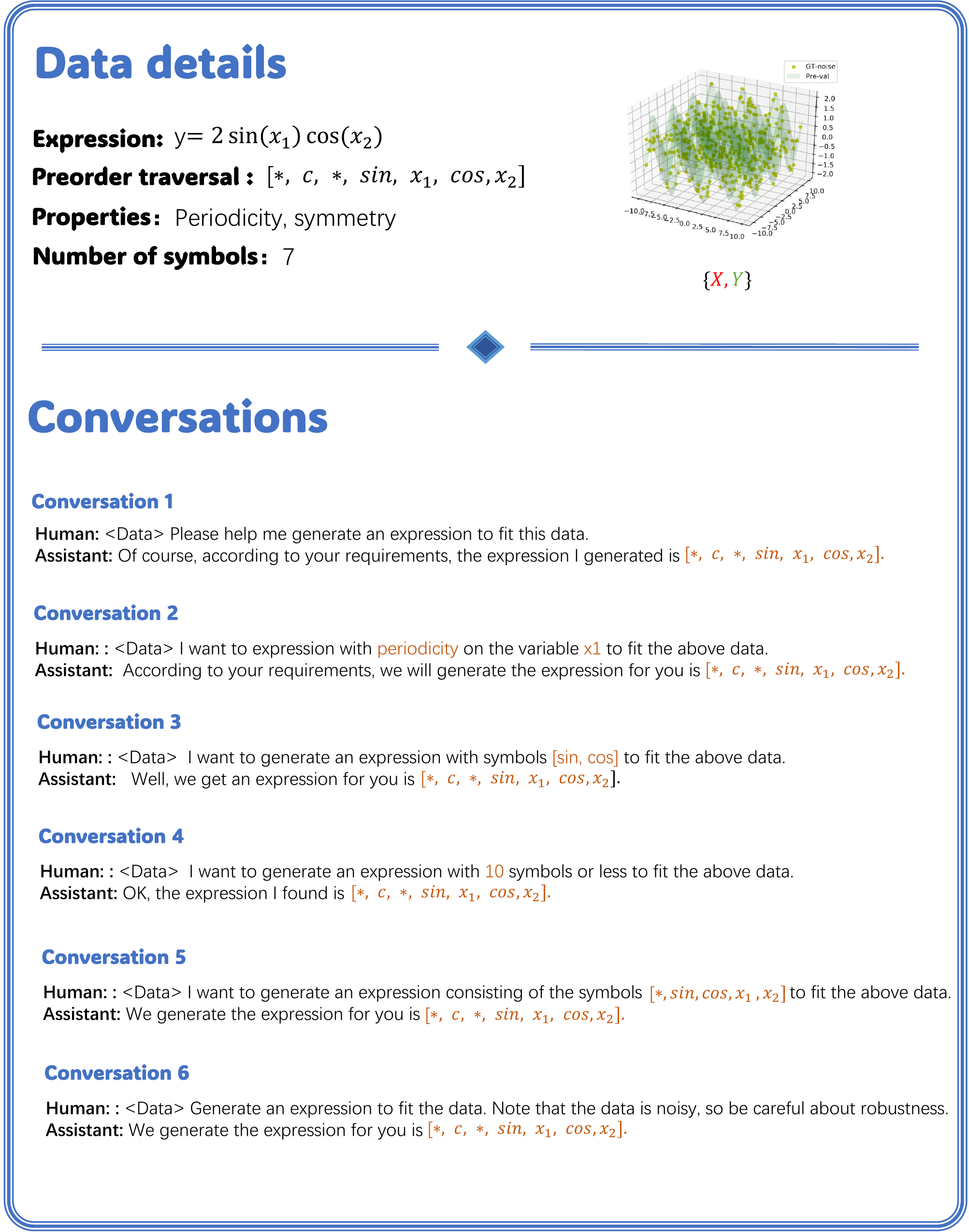}
\caption{Data details and templates. 
Schematic representation of the expression $y=2\sin(x_1)\cos(x_1)$ related question-answer pair generation.
}
\label{fig-converation}
\end{figure*}

In Section~\ref{section3.1}, we described how to generate a large set of expressions. For each expression, we can construct multiple Q\&A-style textual training instances according to specific rules. Concretely, for each expression we have its preorder traversal, several properties (such as periodicity and symmetry), and the length of its preorder sequence. Using this information, we generate question--answer pairs as illustrated in Fig.~\ref{fig-converation}. 

For example, for the expression $y = \sin(x)$, we can construct the following observational data and dialogue pair:
\textbf{\{[X, y]}; \textbf{Human:} \texttt{<Data>} Please generate an expression that fits the uploaded data. \textbf{Assistant:} Of course. According to your request, the expression I generate is $[*, C, *, \sin, x_1, \cos, x_2]$. \textbf{\}}.
Here, $[X, y]$ denotes the observed data; the \textbf{Human} turn represents a natural-language query that may encode prior knowledge; and the \textbf{Assistant} turn corresponds to the response produced by \textsc{ChatSR}. We prepend a special token \texttt{<Data>} to each request sentence to indicate that the following content corresponds to data features. A more detailed explanation of Fig.~\ref{fig-converation} is provided below.

\begin{itemize}
\item[1)] We construct Q\&A pairs that simply ask the model to generate an expression that fits the given data, analogous to traditional symbolic regression.
\item[2)] We construct Q\&A pairs that require the generated expression to satisfy certain properties. For example, the expression $y = 2\sin(x_1)\cos(x_2)$ is periodic with respect to both variables. Based on this, we can ask the model to generate expressions that are periodic in $x_1$, periodic in $x_2$, or periodic in each variable simultaneously. The same strategy is applied to properties such as symmetry.
\item[3)] To enforce that certain symbols must appear in the expression generated by \textsc{ChatSR}, we randomly select $k$ symbols from the preorder traversal $S$ of the ground-truth expression, where $k < |S|$ is a random integer between $1$ and $|S| - 1$. The selected symbols are then inserted into a natural-language template to form a prompt that explicitly requires the model to use these symbols.
\item[4)] We also construct Q\&A pairs that constrain the length of the generated expression. Specifically, we require the length of the preorder traversal of the expression produced by \textsc{ChatSR} to be less than a threshold $M_L$. We obtain $M_L$ by adding a random integer between $0$ and $20$ to the true preorder length of the expression, and then embed $M_L$ into a corresponding natural-language instruction.
\item[5)] In some prompts, we require \textsc{ChatSR} to generate expressions using only a specified subset of symbols. To this end, we process the preorder traversal of the ground-truth expression to remove duplicate symbols, and insert the resulting symbol set into a template sentence to obtain a complete dialogue turn.
\item[6)] To encourage robustness to noise, we add Gaussian noise of varying levels to the clean observational data and then construct Q\&A pairs that ask the model to generate expressions that fit the noisy data while remaining robust.
\end{itemize}

\subsubsection{Multi-turn dialogues}
Multiple-turn dialogues allow us to interact more deeply with the model. To construct multi-turn training data, we proceed as follows.

\textbf{(1)} For each expression and its corresponding observational data generated in Section~\ref{section3.1}, we feed them into DSO for iterative optimization. At each iteration, we record the expression with the highest reward in that round and its reward value $R$. After the inference process terminates, we obtain a chain of inference (COI). For example, for the target expression $y = \sin(x) + \cos(x)$, a possible COI is

\[
\left\{
\begin{aligned}
&[\sin, x; R = 0.2] \rightarrow [+, \sin, x, x; R = 0.6] \rightarrow \dots \\
&\rightarrow [*, \cos, x, x; R = 0.5] \rightarrow [+, \sin, x, \cos, x; R = 1.0]
\end{aligned}
\right\}.
\]

\textbf{(2)} We then sort the expressions in the COI in ascending order of their reward values:
\[
\left\{
\begin{aligned}
&[\sin, x; R = 0.2] \rightarrow [*, \cos, x, x; R = 0.5] \rightarrow \dots \\
&\rightarrow [+, \sin, x, x; R = 0.6] \rightarrow [+, \sin, x, \cos, x; R = 1.0]
\end{aligned}
\right\}.
\]

\textbf{(3)} We select the top $T$ expressions with the largest rewards to form the Optimal Chain of Inference (OCOI). For the above example, if we choose $T = 2$, the resulting OCOI is
\[
\{\,[+, \sin, x, x; R = 0.6] \rightarrow [+, \sin, x, \cos, x; R = 1.0]\,\}.
\]

\textbf{(4)} Given the OCOI, and using the templates and single-turn dialogue construction method described in Section~3.1.1, we can generate multi-turn dialogues. For instance, for the OCOI
\[
\{\,[+, \sin, x, x; R = 0.6] \rightarrow [+, \sin, x, \cos, x; R = 1.0]\,\},
\]
we can obtain the following two-turn dialogue:
\begin{quote}
\textbf{Human:} \texttt{<Data>} Please generate an expression that fits the uploaded data.\\
\textbf{Assistant:} Of course. According to your request, the expression I generate is $[+, \sin, x, x]$.\\[0.3em]
\textbf{Human:} I would like an expression that is periodic in the variable $x$ to fit the above data.\\
\textbf{Assistant:} According to your requirement, we generate the expression $[+, \sin, x, \cos, x]$ for you.
\end{quote}

\subsection{Model architecture}
In \textsc{ChatSR} (Fig.~\ref{fig1}), we employ a SetTransformer trained with contrastive learning as the data feature encoder, and keep its parameters frozen throughout training. The features extracted by the SetTransformer are then mapped into the word-embedding space of the LLM via a projection layer, enabling the language model to interpret the input data. Finally, we jointly train the parameters of the projection layer and the LLM. We choose Qwen3~\cite{qwen3} as our LLM, denoted by $f_{\varphi}$ and parameterized by $\varphi$.

\subsubsection{SetTransformer}
Data information plays a crucial role in guiding the decoder. To respect the permutation invariance of the data---that is, the fact that the representation should not depend on the ordering of input samples---we adopt the SetTransformer as our data encoder, following~\cite{pmlr-v97-lee19d}. Our encoder takes as input a set of data points $\mathcal{D} = \{X, y\}$, where $X \in \mathbb{R}^{n \times d}$ denotes the input features and $y \in \mathbb{R}^n$ denotes the corresponding targets.

The data points are first passed through a trainable affine layer, which projects them into a latent space $h_n \in \mathbb{R}^{d_h}$. The resulting representations are then processed by a stack of Induced Set Attention Blocks (ISABs)~\cite{pmlr-v97-lee19d}, each of which employs cross-attention. In the first cross-attention layer, a set of learnable inducing points is used as queries, while the input data serves as keys and values. The outputs of this layer are then used as keys and values in a subsequent cross-attention layer, with the original data representations acting as queries.

After these attention layers, we apply dropout to mitigate overfitting. Finally, we standardize the output size via a last cross-attention layer, which uses another set of learnable vectors as queries, ensuring that the encoder output has a fixed dimensionality independent of the number of input samples. The hyperparameter settings of the SetTransformer are summarized in Table~\ref{set-tab}.

\label{A1}
\begin{table}[htbp]
\centering
{
\begin{tabular}{lc}
\textbf{hyperparameters} & \textbf{Numerical value}\\ 
\toprule
 \textbf{N\_p}  & 0\\
 \textbf{activation}  & 'relu'\\
 \textbf{bit16}  & True\\
 \textbf{dec\_layers}  & 5\\
 \textbf{dec\_pf\_dim}  & 512\\
 \textbf{dim\_hidden}  & 512\\
 \textbf{dim\_input}  & 3\\
 \textbf{dropout}  & 0\\
 \textbf{input\_normalization}  & False\\
 \textbf{length\_eq}  & 60\\
 \textbf{linear}  & False\\
 \textbf{ln}  & True\\
 \textbf{lr}  & 0.0001\\
 \textbf{mean}  & 0.5\\
 \textbf{n\_l\_enc}  & 5\\
 \textbf{norm}  & True\\
 \textbf{num\_features}  & 20\\
 \textbf{num\_heads}  & 8\\
 \textbf{num\_inds}  & 50\\
 \textbf{output\_dim}  & 60\\
 \textbf{sinuisodal\_embeddings} & False\\
 \textbf{src\_pad\_idx} & 0\\
 \textbf{std} & 0.5\\
 \textbf{trg\_pad\_idx} & 0\\
 \hline
 \end{tabular} 
\caption{Hyperparameters of SetTransformer}
\label{set-tab}
}
\end{table}
\subsubsection{Qwen3}
Qwen3~\cite{qwen3} is a conversational large language model obtained by supervised instruction tuning on the base models of the Tongyi family. Its training corpus consists of high-quality, multi-source instruction data and multi-turn dialogues, covering both Chinese and English, as well as code and retrieval/tool-use scenarios. The model is trained end-to-end with a standard autoregressive objective.

To enhance its ability to model long contexts, Qwen3 supports extended sequence lengths in both the pre-training and fine-tuning stages, and leverages engineering techniques such as gradient checkpointing and flash attention to reduce GPU memory usage and training latency. In the inference stage, Group Query Attention (GQA) and an efficient KV cache are employed to alleviate memory and throughput bottlenecks.

For dialogue alignment, Qwen3 adopts a loss design tailored to multi-turn interaction, updating weights primarily based on the assistant’s outputs, and further applies preference optimization to improve instruction following and contextual coherence. In addition, Qwen3 is specifically tuned for function calling and structured output, yielding more stable interaction performance in complex task orchestration and tool-chain collaboration scenarios.

\subsection{Model training}

For each data instance $X_D = (X, y)$, we construct a multi-turn dialogue consisting of question--answer pairs
\[
[X_q^1, X_a^1, X_q^2, X_a^2, \dots, X_q^T, X_a^T],
\]
where $T$ denotes the total number of dialogue turns. We organize these into a single sequence
\[
[X_D, X_q^1, X_a^1, X_q^2, X_a^2, \dots, X_q^T, X_a^T],
\]
treating all answers as the assistant’s responses. The instruction $X_{\text{instruct}}^t$ at the $t^{\text{th}}$ turn is defined as follows:

\begin{table*}[htp]
\renewcommand{\arraystretch}{1.1}
\centering
\resizebox{18.4cm}{!}{
\def\arraystretch{1.1}
    \small
        \bgroup
        \setlength{\tabcolsep}{0.4em}
        \begin{tabular}{c|l|cccccccccc}
\toprule

            \multirow{2}{*}{\bf Group} & \multicolumn{1}{c|}{\multirow{2}{*}{\bf Dataset}} & \multicolumn{2}{c}{\bf ChatSR} & \multicolumn{2}{c}{\bf MMSR} & \multicolumn{2}{c}{\bf SNIP} & \multicolumn{2}{c}{\bf NeSymRes} & \multicolumn{2}{c}{\bf TPSR }  \\
            \cline{3-12}
& & $R^2 \uparrow$ &Nodes $\downarrow$ & $R^2 \uparrow$ &Nodes $\downarrow$& $R^2 \uparrow$ &Nodes $\downarrow$& $R^2 \uparrow$ &Nodes $\downarrow$& $R^2 \uparrow$ &Nodes $\downarrow$ \\ 
\cmidrule(lr){2-2}
\cmidrule(lr){3-4}
\cmidrule(lr){5-6}
\cmidrule(lr){7-8}
\cmidrule(lr){9-10}
\cmidrule(lr){11-12}
\multirow{10}{*}{\rotatebox{90}{Standards}}
& Nguyen    & $\textbf{0.9999}_{\pm0.001}$&\textbf{12.1} & $\textbf{0.9999}_{\pm0.001}$ &14.8& $0.9945_{\pm0.004}$&18.8& $0.8763_{\pm0.003}$&16.6& $0.9903_{\pm0.004}$ &37.4 \\
& Keijzer   & $\textbf{0.9992}_{\pm0.003}$ &\textbf{13.4}& $0.9924_{\pm0.003}$&17.9& $0.9895_{\pm0.005}$&19.5 & $0.7992_{\pm0.005}$&23.2 & $0.9889_{\pm0.003}$  &35.7\\
& Korns     & $\textbf{0.9941}_{\pm0.003}$ &\textbf{16.4} &$ 0.9927_{\pm0.003}$&19.9& $0.9557_{\pm0.006}$&22.5  & $0.8313_{\pm0.006}$ &23.8& $0.9329_{\pm0.005}$  &38.3\\
& Constant  & $0.9925_{\pm0.002}$&\textbf{20.5}  & $\textbf{0.9946}_{\pm0.002}$ &26.2& $0.9299_{\pm0.005}$ &24.8 & $0.8246_{\pm0.004}$  &25.0& $0.9425_{\pm0.007}$  &44.9 \\
& Livermore  & $\textbf{0.9885}_{\pm0.003}$&\textbf{23.6} & $0.9726_{\pm0.004}$&28.7& $0.9036_{\pm0.003}$&34.6& $0.6692_{\pm0.005}$&34.2 & $0.9014_{\pm0.007}$ &60.3\\
& Vladislavleva  & $\textbf{0.9884}_{\pm0.003}$&\textbf{16.8} & $0.9812_{\pm0.003}$&22.4 & $0.9415_{\pm0.006}$&42.1 & $0.6773_{\pm0.006}$ &37.4& $0.9623_{\pm0.006}$  &68.8\\
& R  & $\textbf{0.9948}_{\pm0.004}$&\textbf{14.3} & $0.9811_{\pm0.004}$&15.8& $0.9583_{\pm0.005}$ &26.7& $0.7663_{\pm0.004}$&28.7 & $0.9253_{\pm0.005}$ &46.3 \\
& Jin  & $\textbf{0.9962}_{\pm0.003}$ &\textbf{21.6}& $0.9902_{\pm0.003}$&31.4 & $0.9877_{\pm0.004}$&17.2 & $0.8246_{\pm0.005}$&21.6& $0.9601_{\pm0.007}$&46.4\\
& Neat  & $0.9943_{\pm0.004}$&\textbf{12.7} & $ \textbf{0.9952}_{\pm0.004}$ &18.8& $ 0.9401_{\pm0.004}$&18.9 & $ 0.7729_{\pm0.005}$&21.4& $ 0.9426_{\pm0.006}$ &39.3\\
& Others  & $0.9936_{\pm0.002}$ &\textbf{15.3} & $\textbf{0.9968}_{\pm0.002}$&23.1& $0.9702_{\pm0.003}$&30.4 & $0.7825_{\pm0.004}$&35.1 & $0.9592_{\pm0.005}$&48.9 \\
\midrule
\multirow{3}{*}{\rotatebox{90}{SRBench}}
& Feynman  & $\textbf{0.9910}_{\pm0.002}$&\textbf{16.4 }&$0.9874_{\pm0.002}$ &22.4&$0.8899_{\pm0.004}$&21.1  & $0.7214_{\pm0.006}$&21.2 &$0.9025_{\pm0.006}$&47.2\\
& Strogatz  & $\textbf{0.9861}_{\pm0.003}$&\textbf{28.7} &$0.9819_{\pm0.003}$&32.1 &$0.8307_{\pm0.003}$&26.8  & $0.6314_{\pm0.005}$&29.3 &$0.8813_{\pm0.006}$&33.5\\
& Black-box & $0.8921_{\pm0.004}$&\textbf{18.9} &$\textbf{0.9037}_{\pm0.004}$ &24.2&$0.8692_{\pm0.004}$ &29.2 & $0.6684_{\pm0.005}$&34.1 &$0.9103_{\pm0.004}$&59.1\\
\cline{2-7} 
\toprule
\multirow{1}{*}{\rotatebox{90}{ }}
 & Average & $\textbf{0.9854}$&\textbf{17.7} &$0.9822$ &22.9&$0.9354$ &25.6 & $0.7573$ &27.0& $0.9384$&46.6\\
 \toprule
        
\end{tabular}
\egroup
}
\caption{The results of performance comparison. At a 0.95 confidence level, a comparison of the coefficient of determination ($R^2$) and the expression complexity(Nodes) was conducted between ChatSR and four baselines.
\label{tab1}}
\end{table*} 
\begin{table*}[htp]
\center
\resizebox{18.4cm}{!}{
\begin{tabular}{c|cccccccccc}
\toprule[1.45pt]
\toprule[1pt]
\cmidrule(lr){2-11}
Points& \multicolumn{2}{c}{Monotonically increasing}& \multicolumn{2}{c}{Monotonically decreasing}&\multicolumn{2}{c}{central symmetry}&\multicolumn{2}{c}{convexity}&\multicolumn{2}{c}{concavity}\\  
\cmidrule(lr){2-11}
Prior-k (Use/No)&Use Prior-k & No Prior-k & Use Prior-k& No Prior-k & Use Prior-k& No Prior-k& Use Prior-k& No Prior-k&Use Prior-k& No Prior-k\\
\cmidrule(lr){1-1}
\cmidrule(lr){2-3}
\cmidrule(lr){4-5}
\cmidrule(lr){6-7}
\cmidrule(lr){8-9}
\cmidrule(lr){10-11}
$R^2$ &  $0.9992 $ & $0.9936$  & $0.9993$&$ 0.9902 $ &$ 0.9953 $&$0.9885$ &$ 0.9999 $&$0.9905$&$ 0.9998 $&$0.9902$\\
Recover rate & $71.8\%$  & $56.2\%$ &$ 73.0\%$ &$ 52.7\%$ &$ 68.4\% $&$44.9\%$&$ 70.2\%$ & $ 48.3\%$ &$ 66.0\% $&$46.7\%$\\
Success rate & $96.2\%$  & $66.2\%$ &$ 97.0\%$ &$ 62.2\%$ &$ 90.4\% $&$38.8\%$&$ 98.5\%$ & $ 88.2\%$ &$ 97.0\% $&$86.2\%$\\
\toprule
\end{tabular}
}
\caption{The Zero-shot proficiency test. In the table, `Prior-k' denotes `Prior knowledge'. Here, `Use prior-k' and `No prior-k' denote whether Prior knowledge of the relevant properties is introduced in the prompt, respectively. The `Success rate' represents the proportion of generated expressions that conform to the corresponding property.
\label{tab_zero_shot}}
\end{table*}
\begin{equation}
X_{\text{instruct}}^t = 
\begin{cases} 
 [X_D, X_q^1], & \text{When t = 1 }\\
X_q^t, & \text{When t \textgreater 1 }
\end{cases}
\end{equation}

We perform instruction-tuning of the LLM on the prediction tokens, using its original auto-regressive training objective.
Specifically, for a sequence of length L, we compute the probability of the target answers $X_a$ by:
\begin{equation}
\label{eq_likehood}
p(X_a \mid X_D, X_{\text{instruct}}) = \prod_{i=1}^{L} p_{\theta}(x_i \mid X_D, X_{\text{instruct}, <i}, X_{a, <i})
\end{equation}

\begin{figure*}[h]
\centering
\vspace{-0.3cm}
\includegraphics[width=180mm]{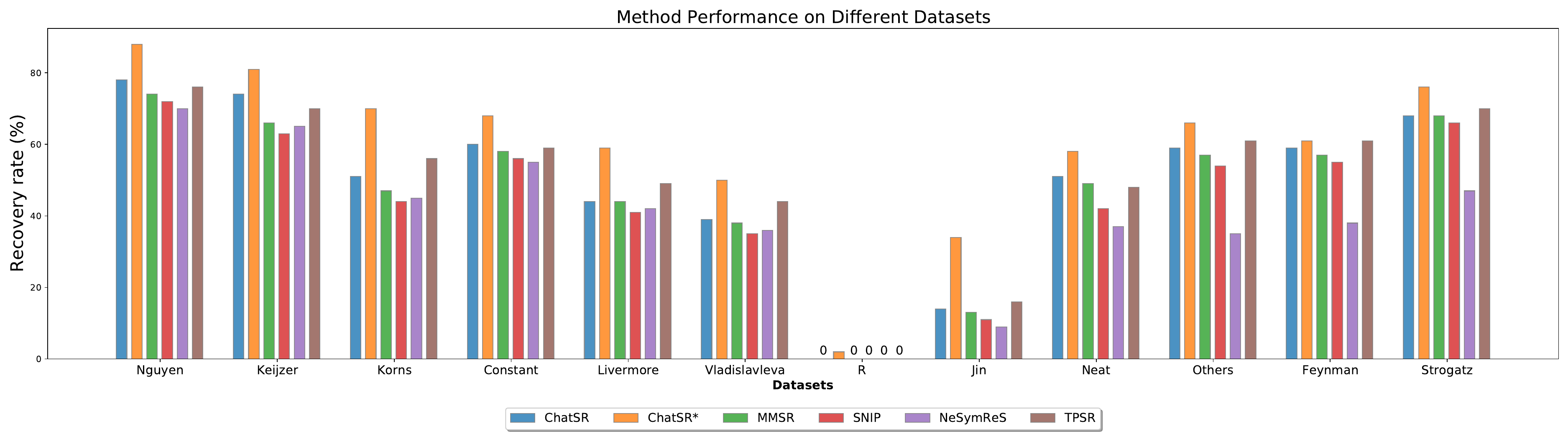}
\caption{Recovery rate of various algorithms.
Note: \textcolor{blue}{ChatSR}'s prompt does not contain prior knowledge. However, \textcolor{orange}{ChatSR*}'s prompt carries prior knowledge. From the figure, we can see that introducing prior knowledge into the instruction can effectively improve the recovery rate.
}
\label{fig-recovery}
\end{figure*}
where $\theta$ denotes the trainable parameters, and $X_{\text{instruct},<i}$ and $X_{a,<i}$ are the instruction and answer tokens, respectively, from all turns preceding the current prediction token $x_i$. We explicitly prepend $X_D$ to the sequence to emphasize that all answers are grounded in the observed data. For training \textsc{ChatSR}, we adopt a two-stage instruction-tuning procedure.

\paragraph{Stage 1: Pre-training for feature alignment}
In the first stage, we sample 600K instances from all datasets (including triples of the form $[X_D, X_q, X_a]$) for feature alignment training. During this stage, we keep both the SetTransformer and the LLM weights frozen, and maximize the likelihood in Eq.~\ref{eq_likehood} using only the trainable parameters $\theta = W$ (the projection matrix). This encourages the data features $H_v$ to be aligned with the pre-trained LLM word-embedding space. Conceptually, this stage can be viewed as training a data tokenizer that is compatible with the frozen LLM.

\paragraph{Stage 2: Fine-tuning the model end-to-end}
In the second stage, we continue to keep the SetTransformer weights frozen, while updating both the projection layer and the LLM within \textsc{ChatSR}. In this stage, the trainable parameters are $\theta = \{W, \varphi\}$ in Fig.~\ref{fig1}.

\subsection{Constant optimization}
The LLM first generates the preorder traversal of an expression. For expressions that contain the constant placeholder $C$, we then apply BFGS or other numerical optimization algorithms to estimate the optimal constant values. Finally, we substitute the optimized constants back into the expression and output the resulting closed-form formula.

For example, suppose the LLM outputs the preorder sequence $[*, C, \sin, x]$, which corresponds to the expression $C \sin(x)$. In this case, we use the BFGS algorithm to optimize the value of $C$ by fitting the expression $C \sin(x)$ to the data, with $X$ as input and $y$ as target.

\section{Experiments}
To evaluate the performance of \textsc{ChatSR}, we conduct experiments on 13 benchmark datasets. We compare \textsc{ChatSR} against four state-of-the-art symbolic regression baselines, summarized below:
\begin{itemize}
\item \textbf{MMSR}~\cite{li2024mmsr}: A pre-training framework that treats symbolic regression as a multimodal learning problem and uses contrastive learning for cross-modal alignment.
\item \textbf{TPSR}~\cite{shojaee2024transformer}: A symbolic regression method that combines large-scale pre-trained transformers with Monte Carlo Tree Search.
\item \textbf{NeSymReS}~\cite{biggio2021neural}: A large-scale pre-trained model designed specifically for symbolic regression.
\item \textbf{SNIP}~\cite{meidani2023snip}: A large-scale pre-trained model that employs a feature encoder trained by contrastive learning prior to symbolic regression training.
\end{itemize}

\subsection{Comparison with baselines without prior knowledge}

\subsubsection{Comparison of $R^2$}
The primary objective of symbolic regression is to discover an analytic expression that accurately fits the observed data. A widely used metric for assessing the goodness of fit is the coefficient of determination, $R^2$, defined as
\begin{equation}
\label{eq:r2}
\mathcal{R}^2 
= 1 - \frac{\sum_{i=1}^{N} (y_i - \hat{y}_i)^2}
              {\sum_{i=1}^{N} (y_i - \overline{y})^2},
\end{equation}
where $\hat{y}_i$ is the predicted value for the $i$-th sample and $\overline{y}$ is the mean of the ground-truth targets $y_i$.

We evaluate all five methods on 13 datasets (see Appendices~\ref{AK} and~\ref{AJ} for details) using $R^2$ as the main metric. For each ground-truth expression in a dataset, we run each method 20 times and compute the average $R^2$ over all expressions. We report results with a confidence level of 0.95~\cite{junk1999confidence,costermans1992confidence}. The detailed results are shown in Table~\ref{tab1} (\(R^2\)).

\subsubsection{Comparison of recovery rate}
Recovery rate is a stricter metric than $R^2$, as it measures how often an algorithm fully recovers the target expression. Intuitively, an expression is considered fully recovered if it is algebraically equivalent to the ground truth and thus achieves, or can achieve, $\mathcal{R}^2 = 1.0$. For example, for the target expression $\sin(x)$, expressions such as $\sin(x) + c$ and $\sin(c x)$ can be regarded as full recoveries under our criterion.

Formally, for each dataset, we compute the recovery rate as the ratio between (i) the total number of successful full recoveries across all runs and (ii) the total number of runs. The average recovery rate for each dataset is reported in Fig.~\ref{fig-recovery}.

\subsubsection{Comparison of complexity (number of nodes)}
The structural complexity of the discovered expression (e.g., the number of nodes in its expression tree) is also an important metric for symbolic regression. Overly complex expressions are less interpretable and often undesirable, even if they achieve high $R^2$. In this experiment, for each algorithm we run 20 trials per ground-truth expression and record both the average $R^2$ and the average number of nodes of the generated expressions. The results are summarized in Table~\ref{tab1} (Nodes).

\subsubsection{Result analysis}
From the above results, we observe that although \textsc{ChatSR} is only slightly better than MMSR in terms of average $\mathcal{R}^2$, it achieves substantially lower expression complexity. We attribute this to the inductive bias of large language models, which tend to favor simpler and more compact expressions. Consequently, \textsc{ChatSR} generates markedly more concise formulas than MMSR and the other baselines.

It is also worth noting that \textsc{ChatSR} outperforms all competing methods in terms of the more challenging recovery rate metric, which we attribute to the strong prior knowledge encoded in the underlying large language model.

\begin{table}[!t]
\center
\fontsize{16pt}{17.6pt}\selectfont
\resizebox{85mm}{!}{
\begin{tabular}{ccccc}
\toprule[1.45pt]
ChatSR & \multicolumn{2}{c}{With prior knowledge(\textcolor{blue}{ChatSR})}& \multicolumn{2}{c}{Without prior knowledge(\textcolor{orange}{ChatSR*})}\\  
\cmidrule[0.1pt]{1-5}
&$R^2 \uparrow$  &Nodes $\downarrow$  & $R^2 \uparrow$ &Nodes $\downarrow$  \\ 
\cmidrule(lr){1-1}
\cmidrule(lr){2-3}
\cmidrule(lr){4-5}
Nguyen         & 0.9999 & 12.1 & 0.9999 & 9.8 \\
Keijzer        & 0.9992 & 13.4 & 0.9996 & 10.6 \\
Korns          & 0.9941 & 16.4 & 0.9965 & 12.1 \\
Constant       & 0.9925 & 20.5 & 0.9953 & 14.9 \\
Livermore      & 0.9885 & 23.6 & 0.9924 & 17.1 \\
Vladislavleva  & 0.9884 & 16.8 & 0.9918 & 12.2 \\
R              & 0.9948 & 14.3 & 0.9972 & 10.5 \\
Jin            & 0.9962 & 21.6 & 0.9981 & 15.8 \\
Neat           & 0.9943 & 12.7 & 0.9967 & 9.4  \\
Others         & 0.9936 & 15.3 & 0.9960 & 11.1 \\
Feynman        & 0.9910 & 16.4 & 0.9941 & 12.3 \\
Strogatz       & 0.9861 & 28.7 & 0.9912 & 21.6 \\
\cmidrule(lr){1-1}
\cmidrule(lr){2-3}
\cmidrule(lr){4-5}
Average        & 0.9932 & 17.7 & 0.9957 & 13.1 \\
\bottomrule

\end{tabular}
}
\caption{Performance comparison of ChatSR before and after introducing prior knowledge. 
\label{pk}}
\end{table}

\subsection{Ablation study: Does prior knowledge improve \textsc{ChatSR}?}

Expression recovery rate is a particularly challenging metric for evaluating symbolic regression algorithms. To examine whether introducing prior knowledge into the prompts can improve the recovery rate of \textsc{ChatSR}, we conduct the following ablation study.

For each dataset, we consider two settings. In the first setting, the prompt does not provide any prior knowledge (i.e., no additional constraints); we simply ask the model to generate an expression that fits the data. In the second setting, we augment the prompt with prior knowledge derived from the properties and structure of the ground-truth expression. For example, for the target expression $\sin(x) + \cos(x)$, we may ask the model to generate an expression that contains the symbols $\sin$ and $\cos$, or to generate an expression that is periodic.

For each expression, we run both settings 20 times. We then compute the average $\mathcal{R}^2$ (Table~\ref{pk}) and the recovery rate (Fig.~\ref{fig-recovery}) over all runs. In the following, we refer to the setting without prior knowledge as \textsc{ChatSR}, and to the setting with prior knowledge in the prompt as \textsc{ChatSR*}.

From Table~\ref{pk} and Fig.~\ref{fig-recovery}, we observe that providing prior knowledge in the prompt significantly improves the recovery rate. This confirms our goal of enhancing the quality of the generated expressions by incorporating priors through natural-language instructions. More importantly, it demonstrates that \textsc{ChatSR} is indeed capable of generating expressions that satisfy user-specified requirements.

\subsection{Zero-shot ability: Can \textsc{ChatSR} understand unseen prior knowledge?}

General-purpose large language models are known to possess strong natural-language understanding and extensive background knowledge, and have shown impressive zero-shot performance on many tasks. We would like \textsc{ChatSR} to inherit this capability as well. For instance, if the training data only encode properties such as symmetry and periodicity, can \textsc{ChatSR} still exploit the zero-shot ability of the underlying LLM to understand and generate expressions with other properties (e.g., monotonicity)?

To test the zero-shot capability of \textsc{ChatSR}, we select several properties that are not included in the training dataset: monotonicity, symmetry with respect to the origin, convexity, concavity, and boundedness. Our goal is to assess whether \textsc{ChatSR} can, purely based on its language understanding, generate expressions that satisfy these requirements when they are described in the prompt.

Concretely, for each of the above properties, we manually synthesize 10 expressions that exhibit the desired behavior (see Appendix~\ref{p-expression}). For each such expression, we run 20 trials under different prompting conditions, specifying whether we require the generated expression to satisfy the given property or not. We then compute the average $\mathcal{R}^2$, the recovery rate, and the success rate, defined as the proportion of generated expressions that actually satisfy the specified property. The results are reported in Table~\ref{tab_zero_shot}.

As shown in Table~\ref{tab_zero_shot}, \textsc{ChatSR} exhibits strong zero-shot ability. When provided with an appropriate natural-language description of the desired property, \textsc{ChatSR} is able to generate expressions that meet the requirement, even though such properties were not present in the training dataset. At the same time, both $\mathcal{R}^2$ and recovery rate are improved. We attribute this behavior to the powerful language understanding capabilities of the underlying large language model.

\section{Conclusion and Discussion}
In this work, we have presented \textsc{ChatSR}, a novel symbolic regression framework based on multimodal large language models that enables users to interact and specify requirements via natural-language dialogue. Concretely, we train a multimodal LLM on a large corpus of data--dialogue pairs so that it learns to fit observational data by generating analytic expressions conditioned on both the data and natural-language prompts. Our experiments further show that \textsc{ChatSR} exhibits strong zero-shot capabilities, allowing it to handle various forms of prior knowledge that were not explicitly seen during training.

This paradigm has the potential to change how symbolic regression is applied in practice. When users require the generated expression to satisfy certain constraints, they can simply describe these requirements in natural language, rather than modifying algorithmic code or designing bespoke search heuristics. This substantially lowers the barrier to using symbolic regression and improves the flexibility and usability of symbolic regression methods.

Because \textsc{ChatSR} produces interpretable mathematical expressions from data, it holds considerable promise for applications in domains such as finance and healthcare, where interpretability is critical. In addition, we believe \textsc{ChatSR} has significant potential in scientific discovery and AI for Science more broadly.

Nonetheless, \textsc{ChatSR} also has limitations. For example, its robustness to noise is still insufficient in some settings. As future work, we plan to improve its noise robustness using contrastive learning and other advanced training strategies.

\bibliographystyle{IEEEtran}
\bibliography{Ref}

@article{petersen2019deep,
  title={Deep symbolic regression: Recovering mathematical expressions from data via risk-seeking policy gradients},
  author={Petersen, Brenden K and Landajuela, Mikel and Mundhenk, T Nathan and Santiago, Claudio P and Kim, Soo K and Kim, Joanne T},
  journal={arXiv preprint arXiv:1912.04871},
  year={2019}
}

@article{mundhenk2021symbolic,
  title={Symbolic regression via neural-guided genetic programming population seeding},
  author={Mundhenk, T Nathan and Landajuela, Mikel and Glatt, Ruben and Santiago, Claudio P and Faissol, Daniel M and Petersen, Brenden K},
  journal={arXiv preprint arXiv:2111.00053},
  year={2021}
}

@article{sun2022symbolic,
  title={Symbolic physics learner: Discovering governing equations via monte carlo tree search},
  author={Sun, Fangzheng and Liu, Yang and Wang, Jian-Xun and Sun, Hao},
  journal={arXiv preprint arXiv:2205.13134},
  year={2022}
}

@article{shojaee2024transformer,
  title={Transformer-based Planning for Symbolic Regression},
  author={Shojaee, Parshin and Meidani, Kazem and Barati Farimani, Amir and Reddy, Chandan},
  journal={Advances in Neural Information Processing Systems},
  volume={36},
  year={2024}
}

@article{kamienny2023deep,
  title={Deep Generative Symbolic Regression with Monte-Carlo-Tree-Search},
  author={Kamienny, Pierre-Alexandre and Lample, Guillaume and Lamprier, Sylvain and Virgolin, Marco},
  journal={arXiv preprint arXiv:2302.11223},
  year={2023}
}

@article{li2024discovering,
  title={Discovering Mathematical Formulas from Data via GPT-guided Monte Carlo Tree Search},
  author={Li, Yanjie and Li, Weijun and Yu, Lina and Wu, Min and Liu, Jingyi and Li, Wenqiang and Hao, Meilan and Wei, Shu and Deng, Yusong},
  journal={arXiv preprint arXiv:2401.14424},
  year={2024}
}

@article{valipour2021symbolicgpt,
  title={Symbolicgpt: A generative transformer model for symbolic regression},
  author={Valipour, Mojtaba and You, Bowen and Panju, Maysum and Ghodsi, Ali},
  journal={arXiv preprint arXiv:2106.14131},
  year={2021}
}

@inproceedings{biggio2021neural,
  title={Neural symbolic regression that scales},
  author={Biggio, Luca and Bendinelli, Tommaso and Neitz, Alexander and Lucchi, Aurelien and Parascandolo, Giambattista},
  booktitle={International Conference on Machine Learning},
  pages={936--945},
  year={2021},
  organization={PMLR}
}

@article{kamienny2022end,
  title={End-to-end symbolic regression with transformers},
  author={Kamienny, Pierre-Alexandre and d'Ascoli, St{\'e}phane and Lample, Guillaume and Charton, Fran{\c{c}}ois},
  journal={Advances in Neural Information Processing Systems},
  volume={35},
  pages={10269--10281},
  year={2022}
}

@article{meidani2023snip,
  title={SNIP: Bridging Mathematical Symbolic and Numeric Realms with Unified Pre-training},
  author={Meidani, Kazem and Shojaee, Parshin and Reddy, Chandan K and Farimani, Amir Barati},
  journal={arXiv preprint arXiv:2310.02227},
  year={2023}
}

@article{kim2020integration,
  title={Integration of neural network-based symbolic regression in deep learning for scientific discovery},
  author={Kim, Samuel and Lu, Peter Y and Mukherjee, Srijon and Gilbert, Michael and Jing, Li and {\v{C}}eperi{\'c}, Vladimir and Solja{\v{c}}i{\'c}, Marin},
  journal={IEEE transactions on neural networks and learning systems},
  volume={32},
  number={9},
  pages={4166--4177},
  year={2020},
  publisher={IEEE}
}

@article{udrescu2020ai,
  title={AI Feynman: A physics-inspired method for symbolic regression},
  author={Udrescu, Silviu-Marian and Tegmark, Max},
  journal={Science Advances},
  volume={6},
  number={16},
  pages={eaay2631},
  year={2020},
  publisher={American Association for the Advancement of Science}
}

@article{li2023metasymnet,
  title={MetaSymNet: A Dynamic Symbolic Regression Network Capable of Evolving into Arbitrary Formulations},
  author={Li, Yanjie and Li, Weijun and Yu, Lina and Wu, Min and Liu, Jinyi and Li, Wenqiang and Hao, Meilan and Wei, Shu and Deng, Yusong},
  journal={arXiv preprint arXiv:2311.07326},
  year={2023}
}

@inproceedings{pmlr-v97-lee19d,
  title = {Set Transformer: A Framework for Attention-based Permutation-Invariant Neural Networks},
  author = {Lee, Juho and Lee, Yoonho and Kim, Jungtaek and Kosiorek, Adam and Choi, Seungjin and Teh, Yee Whye},
  booktitle = {Proceedings of the 36th International Conference on Machine Learning},
  pages = {3744--3753},
  year = 	{2019},
  volume = {97},
  series = {Proceedings of Machine Learning Research},
  month = {09--15 Jun},
  publisher =  {PMLR},
  url = {https://proceedings.mlr.press/v97/lee19d.html}
}

@article{bojanowski2017optimizing,
  title={Optimizing the latent space of generative networks},
  author={Bojanowski, Piotr and Joulin, Armand and Lopez-Paz, David and Szlam, Arthur},
  journal={arXiv preprint arXiv:1707.05776},
  year={2017}
}

@inproceedings{10.1145/2576768.2598291,
    author = {Arnaldo, Ignacio and Krawiec, Krzysztof and O'Reilly, Una-May},
    title = {Multiple Regression Genetic Programming},
    year = {2014},
    isbn = {9781450326629},
    publisher = {Association for Computing Machinery},
    address = {New York, NY, USA},
    url = {https://doi.org/10.1145/2576768.2598291},
    doi = {10.1145/2576768.2598291},
    booktitle = {Proceedings of the 2014 Annual Conference on Genetic and Evolutionary Computation},
    pages = {879--886},
    numpages = {8}
}

@book{McConaghy2011,
    author={McConaghy, Trent},
    title={FFX: Fast, Scalable, Deterministic Symbolic Regression Technology},
    bookTitle={Genetic Programming Theory and Practice IX},
    year={2011},
    publisher={Springer New York},
    address={New York, NY},
    pages={235--260},
    isbn={978-1-4614-1770-5},
    doi={10.1007/978-1-4614-1770-5\_13},
    url={https://doi.org/10.1007/978-1-4614-1770-5\_13}
}

@article{7473913,
  author={Nguyen, Su and Zhang, Mengjie and Tan, Kay Chen},
  journal={IEEE Transactions on Cybernetics}, 
  title={Surrogate-Assisted Genetic Programming With Simplified Models for Automated Design of Dispatching Rules}, 
  year={2017},
  volume={47},
  number={9},
  pages={2951-2965},
  doi={10.1109/TCYB.2016.2562674}}

@article{xureinforcement,
  title={Reinforcement Symbolic Regression Machine},
  author={Xu, Yilong and Liu, Yang and Sun, Hao}
}

@article{hasselt2010double,
  title={Double Q-learning},
  author={Hasselt, Hado},
  journal={Advances in neural information processing systems},
  volume={23},
  year={2010}
}

@inproceedings{coulom2006efficient,
  title={Efficient selectivity and backup operators in Monte-Carlo tree search},
  author={Coulom, R{\'e}mi},
  booktitle={International conference on computers and games},
  pages={72--83},
  year={2006},
  organization={Springer}
}

@inproceedings{fong2022rethinking,
  title={Rethinking symbolic regression: Morphology and adaptability in the context of evolutionary algorithms},
  author={Fong, Kei Sen and Wongso, Shelvia and Motani, Mehul},
  booktitle={The Eleventh International Conference on Learning Representations},
  year={2022}
}

@article{liu1989limited,
  title={On the limited memory BFGS method for large scale optimization},
  author={Liu, Dong C and Nocedal, Jorge},
  journal={Mathematical programming},
  volume={45},
  number={1-3},
  pages={503--528},
  year={1989},
  publisher={Springer}
}

@article{junk1999confidence,
  title={Confidence level computation for combining searches with small statistics},
  author={Junk, Thomas},
  journal={Nuclear Instruments and Methods in Physics Research Section A: Accelerators, Spectrometers, Detectors and Associated Equipment},
  volume={434},
  number={2-3},
  pages={435--443},
  year={1999},
  publisher={Elsevier}
}

@article{costermans1992confidence,
  title={Confidence level and feeling of knowing in question answering: The weight of inferential processes.},
  author={Costermans, Jean and Lories, Guy and Ansay, Catherine},
  journal={Journal of Experimental Psychology: Learning, Memory, and Cognition},
  volume={18},
  number={1},
  pages={142},
  year={1992},
  publisher={American Psychological Association}
}

@article{vastl2024symformer,
  title={Symformer: End-to-end symbolic regression using transformer-based architecture},
  author={Vastl, Martin and Kulh{\'a}nek, Jon{\'a}{\v{s}} and Kubal{\'\i}k, Ji{\v{r}}{\'\i} and Derner, Erik and Babu{\v{s}}ka, Robert},
  journal={IEEE Access},
  year={2024},
  publisher={IEEE}
}

@article{li2024mmsr,
  title={MMSR: Symbolic regression is a multi-modal information fusion task},
  author={Li, Yanjie and Liu, Jingyi and Wu, Min and Yu, Lina and Li, Weijun and Ning, Xin and Li, Wenqiang and Hao, Meilan and Deng, Yusong and Wei, Shu},
  journal={Information Fusion},
  pages={102681},
  year={2024},
  publisher={Elsevier}
}

@inproceedings{bendinelli2023controllable,
  title={Controllable neural symbolic regression},
  author={Bendinelli, Tommaso and Biggio, Luca and Kamienny, Pierre-Alexandre},
  booktitle={International Conference on Machine Learning},
  pages={2063--2077},
  year={2023},
  organization={PMLR}
}

@inproceedings{CLIP,
  title={Learning transferable visual models from natural language supervision},
  author={Radford, Alec and Kim, Jong Wook and Hallacy, Chris and Ramesh, Aditya and Goh, Gabriel and Agarwal, Sandhini and Sastry, Girish and Askell, Amanda and Mishkin, Pamela and Clark, Jack and others},
  booktitle={International conference on machine learning},
  pages={8748--8763},
  year={2021},
  organization={PMLR}
}

@inproceedings{ALIGN,
  title={Scaling up visual and vision-language representation learning with noisy text supervision},
  author={Jia, Chao and Yang, Yinfei and Xia, Ye and Chen, Yi-Ting and Parekh, Zarana and Pham, Hieu and Le, Quoc and Sung, Yun-Hsuan and Li, Zhen and Duerig, Tom},
  booktitle={International conference on machine learning},
  pages={4904--4916},
  year={2021},
  organization={PMLR}
}

@inproceedings{he2020momentum,
  title={Momentum contrast for unsupervised visual representation learning},
  author={He, Kaiming and Fan, Haoqi and Wu, Yuxin and Xie, Saining and Girshick, Ross},
  booktitle={Proceedings of the IEEE/CVF conference on computer vision and pattern recognition},
  pages={9729--9738},
  year={2020}
}

@inproceedings{chen2020simple,
  title={A simple framework for contrastive learning of visual representations},
  author={Chen, Ting and Kornblith, Simon and Norouzi, Mohammad and Hinton, Geoffrey},
  booktitle={International conference on machine learning},
  pages={1597--1607},
  year={2020},
  organization={PMLR}
}

@article{li2020prototypical,
  title={Prototypical contrastive learning of unsupervised representations},
  author={Li, Junnan and Zhou, Pan and Xiong, Caiming and Hoi, Steven CH},
  journal={arXiv preprint arXiv:2005.04966},
  year={2020}
}

@article{li2020mopro,
  title={Mopro: Webly supervised learning with momentum prototypes},
  author={Li, Junnan and Xiong, Caiming and Hoi, Steven CH},
  journal={arXiv preprint arXiv:2009.07995},
  year={2020}
}

@inproceedings{kim2021vilt,
  title={Vilt: Vision-and-language transformer without convolution or region supervision},
  author={Kim, Wonjae and Son, Bokyung and Kim, Ildoo},
  booktitle={International Conference on Machine Learning},
  pages={5583--5594},
  year={2021},
  organization={PMLR}
}

@inproceedings{antol2015vqa,
  title={Vqa: Visual question answering},
  author={Antol, Stanislaw and Agrawal, Aishwarya and Lu, Jiasen and Mitchell, Margaret and Batra, Dhruv and Zitnick, C Lawrence and Parikh, Devi},
  booktitle={Proceedings of the IEEE international conference on computer vision},
  pages={2425--2433},
  year={2015}
}

@article{wang2021simvlm,
  title={Simvlm: Simple visual language model pretraining with weak supervision},
  author={Wang, Zirui and Yu, Jiahui and Yu, Adams Wei and Dai, Zihang and Tsvetkov, Yulia and Cao, Yuan},
  journal={arXiv preprint arXiv:2108.10904},
  year={2021}
}

@inproceedings{wang2022ofa,
  title={Ofa: Unifying architectures, tasks, and modalities through a simple sequence-to-sequence learning framework},
  author={Wang, Peng and Yang, An and Men, Rui and Lin, Junyang and Bai, Shuai and Li, Zhikang and Ma, Jianxin and Zhou, Chang and Zhou, Jingren and Yang, Hongxia},
  booktitle={International Conference on Machine Learning},
  pages={23318--23340},
  year={2022},
  organization={PMLR}
}

@article{piergiovanni2022answer,
  title={Answer-me: Multi-task open-vocabulary visual question answering},
  author={Piergiovanni, AJ and Li, Wei and Kuo, Weicheng and Saffar, Mohammad and Bertsch, Fred and Angelova, Anelia},
  journal={arXiv preprint arXiv:2205.00949},
  year={2022}
}

@inproceedings{singh2022flava,
  title={Flava: A foundational language and vision alignment model},
  author={Singh, Amanpreet and Hu, Ronghang and Goswami, Vedanuj and Couairon, Guillaume and Galuba, Wojciech and Rohrbach, Marcus and Kiela, Douwe},
  booktitle={Proceedings of the IEEE/CVF Conference on Computer Vision and Pattern Recognition},
  pages={15638--15650},
  year={2022}
}

@article{li2021align,
  title={Align before fuse: Vision and language representation learning with momentum distillation},
  author={Li, Junnan and Selvaraju, Ramprasaath and Gotmare, Akhilesh and Joty, Shafiq and Xiong, Caiming and Hoi, Steven Chu Hong},
  journal={Advances in neural information processing systems},
  volume={34},
  pages={9694--9705},
  year={2021}
}

@inproceedings{li2022blip,
  title={Blip: Bootstrapping language-image pre-training for unified vision-language understanding and generation},
  author={Li, Junnan and Li, Dongxu and Xiong, Caiming and Hoi, Steven},
  booktitle={International Conference on Machine Learning},
  pages={12888--12900},
  year={2022},
  organization={PMLR}
}

@article{chen2023pali,
  title={PaLI-X: On Scaling up a Multilingual Vision and Language Model},
  author={Chen, Xi and Djolonga, Josip and Padlewski, Piotr and Mustafa, Basil and Changpinyo, Soravit and Wu, Jialin and Ruiz, Carlos Riquelme and Goodman, Sebastian and Wang, Xiao and Tay, Yi and others},
  journal={arXiv preprint arXiv:2305.18565},
  year={2023}
}

@article{liu2024visual,
  title={Visual instruction tuning},
  author={Liu, Haotian and Li, Chunyuan and Wu, Qingyang and Lee, Yong Jae},
  journal={Advances in neural information processing systems},
  volume={36},
  year={2024}
}

@article{beit3,
  title={Image as a foreign language: Beit pretraining for all vision and vision-language tasks},
  author={Wang, Wenhui and Bao, Hangbo and Dong, Li and Bjorck, Johan and Peng, Zhiliang and Liu, Qiang and Aggarwal, Kriti and Mohammed, Owais Khan and Singhal, Saksham and Som, Subhojit and others},
  journal={arXiv preprint arXiv:2208.10442},
  year={2022}
}

@article{llava,
  title={Visual instruction tuning},
  author={Liu, Haotian and Li, Chunyuan and Wu, Qingyang and Lee, Yong Jae},
  journal={Advances in neural information processing systems},
  volume={36},
  year={2024}
}

@article{llm1,
  title={A survey on evaluation of large language models},
  author={Chang, Yupeng and Wang, Xu and Wang, Jindong and Wu, Yuan and Yang, Linyi and Zhu, Kaijie and Chen, Hao and Yi, Xiaoyuan and Wang, Cunxiang and Wang, Yidong and others},
  journal={ACM Transactions on Intelligent Systems and Technology},
  year={2023},
  publisher={ACM New York, NY}
}

@article{llm2,
  title={A survey of large language models},
  author={Zhao, Wayne Xin and Zhou, Kun and Li, Junyi and Tang, Tianyi and Wang, Xiaolei and Hou, Yupeng and Min, Yingqian and Zhang, Beichen and Zhang, Junjie and Dong, Zican and others},
  journal={arXiv preprint arXiv:2303.18223},
  year={2023}
}

@article{llm3,
  title={Llama: Open and efficient foundation language models},
  author={Touvron, Hugo and Lavril, Thibaut and Izacard, Gautier and Martinet, Xavier and Lachaux, Marie-Anne and Lacroix, Timoth{\'e}e and Rozi{\`e}re, Baptiste and Goyal, Naman and Hambro, Eric and Azhar, Faisal and others},
  journal={arXiv preprint arXiv:2302.13971},
  year={2023}
}

@article{llm4,
  title={Glm-130b: An open bilingual pre-trained model},
  author={Zeng, Aohan and Liu, Xiao and Du, Zhengxiao and Wang, Zihan and Lai, Hanyu and Ding, Ming and Yang, Zhuoyi and Xu, Yifan and Zheng, Wendi and Xia, Xiao and others},
  journal={arXiv preprint arXiv:2210.02414},
  year={2022}
}

@article{llm5,
  title={Training language models to follow instructions with human feedback},
  author={Ouyang, Long and Wu, Jeffrey and Jiang, Xu and Almeida, Diogo and Wainwright, Carroll and Mishkin, Pamela and Zhang, Chong and Agarwal, Sandhini and Slama, Katarina and Ray, Alex and others},
  journal={Advances in neural information processing systems},
  volume={35},
  pages={27730--27744},
  year={2022}
}

@article{llmsr1,
  title={Llm-sr: Scientific equation discovery via programming with large language models},
  author={Shojaee, Parshin and Meidani, Kazem and Gupta, Shashank and Farimani, Amir Barati and Reddy, Chandan K},
  journal={arXiv preprint arXiv:2404.18400},
  year={2024}
}

@article{llmsr2,
  title={In-Context Symbolic Regression: Leveraging Language Models for Function Discovery},
  author={Merler, Matteo and Dainese, Nicola and Haitsiukevich, Katsiaryna},
  journal={arXiv preprint arXiv:2404.19094},
  year={2024}
}

@article{yu2022coca,
  title={Coca: Contrastive captioners are image-text foundation models},
  author={Yu, Jiahui and Wang, Zirui and Vasudevan, Vijay and Yeung, Legg and Seyedhosseini, Mojtaba and Wu, Yonghui},
  journal={arXiv preprint arXiv:2205.01917},
  year={2022}
}

@article{qwen2.5-vl,
  title={Qwen2. 5-vl technical report},
  author={Bai, Shuai and Chen, Keqin and Liu, Xuejing and Wang, Jialin and Ge, Wenbin and Song, Sibo and Dang, Kai and Wang, Peng and Wang, Shijie and Tang, Jun and others},
  journal={arXiv preprint arXiv:2502.13923},
  year={2025}
}

@article{qwen3,
  title={Qwen3 technical report},
  author={Yang, An and Li, Anfeng and Yang, Baosong and Zhang, Beichen and Hui, Binyuan and Zheng, Bo and Yu, Bowen and Gao, Chang and Huang, Chengen and Lv, Chenxu and others},
  journal={arXiv preprint arXiv:2505.09388},
  year={2025}
}

@article{dai2023instructblip,
  title={Instructblip: Towards general-purpose vision-language models with instruction tuning},
  author={Dai, Wenliang and Li, Junnan and Li, Dongxu and Tiong, Anthony and Zhao, Junqi and Wang, Weisheng and Li, Boyang and Fung, Pascale N and Hoi, Steven},
  journal={Advances in neural information processing systems},
  volume={36},
  pages={49250--49267},
  year={2023}
}

@article{li2024llava,
  title={Llava-next-interleave: Tackling multi-image, video, and 3d in large multimodal models},
  author={Li, Feng and Zhang, Renrui and Zhang, Hao and Zhang, Yuanhan and Li, Bo and Li, Wei and Ma, Zejun and Li, Chunyuan},
  journal={arXiv preprint arXiv:2407.07895},
  year={2024}
}

@article{paligemma1,
  title={Paligemma: A versatile 3b vlm for transfer},
  author={Beyer, Lucas and Steiner, Andreas and Pinto, Andr{\'e} Susano and Kolesnikov, Alexander and Wang, Xiao and Salz, Daniel and Neumann, Maxim and Alabdulmohsin, Ibrahim and Tschannen, Michael and Bugliarello, Emanuele and others},
  journal={arXiv preprint arXiv:2407.07726},
  year={2024}
}

@article{paligemma2,
  title={Paligemma 2: A family of versatile vlms for transfer},
  author={Steiner, Andreas and Pinto, Andr{\'e} Susano and Tschannen, Michael and Keysers, Daniel and Wang, Xiao and Bitton, Yonatan and Gritsenko, Alexey and Minderer, Matthias and Sherbondy, Anthony and Long, Shangbang and others},
  journal={arXiv preprint arXiv:2412.03555},
  year={2024}
}

@article{qwen2-vl,
  title={Qwen2-vl: Enhancing vision-language model's perception of the world at any resolution},
  author={Wang, Peng and Bai, Shuai and Tan, Sinan and Wang, Shijie and Fan, Zhihao and Bai, Jinze and Chen, Keqin and Liu, Xuejing and Wang, Jialin and Ge, Wenbin and others},
  journal={arXiv preprint arXiv:2409.12191},
  year={2024}
}

@article{qwen25-vl,
  title={Qwen2. 5-vl technical report},
  author={Bai, Shuai and Chen, Keqin and Liu, Xuejing and Wang, Jialin and Ge, Wenbin and Song, Sibo and Dang, Kai and Wang, Peng and Wang, Shijie and Tang, Jun and others},
  journal={arXiv preprint arXiv:2502.13923},
  year={2025}
}

\newpage

\appendices

\section{Appendix: Expression details of $Knowledge$}
\label{p-expression}

The dataset contains five properties common to mathematical expressions (Continuous Monotonic Decreasing, Continuous Globally Monotonically Increasing, Origin-Centered Symmetric, Complex Continuous Convex, and Complex Continuous Concave), with 10 expressions collected for each property. They can be used to test the ability of symbolic regression algorithms to understand prior knowledge.
\begin{table}[h!]
\centering
\renewcommand{\arraystretch}{1.2}  
\setlength{\tabcolsep}{12pt}  
\begin{tabular}{c p{32mm} c}
\toprule
\textbf{Function Index} & \textbf{Expression} & \textbf{Domain} \\
\midrule
1  & $f(x) = -x - \ln(x+1)$ & $x \geq 0$ \\
2  & $f(x) = e^{-x} - x^2$ & \( x \in \mathbb{R} \) \\
3  & $f(x) = \frac{1}{x+1} - \sqrt{x}$ & $x > 0$ \\
4  & $f(x) = 10 - x^2 - \arctan(x)$ & \( x \in \mathbb{R} \) \\
5  & $f(x) = \frac{1}{\sqrt{x+1}} - \ln(x+2)$ & $x \geq 0$ \\
6  & $f(x) = e^{-x} \cos(x) + \frac{1}{x+1}$ & $x > 0$ \\
7  & $f(x) = -\ln(x+1) + x^{-0.5}$ & $x > 0$ \\
8  & $f(x) = \sqrt{x+1} - 3\ln(x+2)$ & $x \geq 0$ \\
9  & $f(x) = e^{-x^2} - x$ & \( x \in \mathbb{R} \) \\
10 & $f(x) = -x^{3/2} - \tan^{-1}(x)$ & $x \geq 0$ \\
\bottomrule
\end{tabular}
\caption{List of Continuous Monotonic Decreasing Functions}
\end{table}

\begin{table}[h!]
\centering
\renewcommand{\arraystretch}{1.0} 
\setlength{\tabcolsep}{12pt} 
\begin{tabular}{c p{32mm} c}
\toprule
\textbf{Function Index} & \textbf{Expression} & \textbf{Domain} \\
\midrule
1 & \( f(x) = -x^4 + 2x^2 + 1 \) & \( x \in \mathbb{R} \) \\
2 & \( f(x) = -e^x + 3x - 2 \) & \( x \in \mathbb{R} \) \\
3 & \( f(x) = -x^2 + \log(1 + x^2) \) & \( x \in \mathbb{R} \)\\
4 & \( f(x) = -\cosh(x) \) & \( x \in \mathbb{R} \) \\
5 & \( f(x) = -x^2 - \sqrt{x^2 + 1} \) & \( x \in \mathbb{R} \) \\
6 & \( f(x) = -e^{x/2} - x^2 \) & \( x \in \mathbb{R} \) \\
7 & \( f(x) = \log(1 + e^{-x}) \) & \( x \in \mathbb{R} \) \\
8 & \( f(x) = -\sqrt{1 + x^2} \) & \( x \in \mathbb{R} \)\\
9 & \( f(x) = -\log(x^2 + 1) \) & \( x \in \mathbb{R} \) \\
10 & \( f(x) = -x^6 - x^4 + x^3 - x + e^{-x^2} \) & \( x \in \mathbb{R} \)\\
\bottomrule
\end{tabular}
\caption{List of Complex Continuous Concave Functions}
\end{table}

\begin{table}[h!]
\centering
\renewcommand{\arraystretch}{1.2}  
\setlength{\tabcolsep}{12pt}  
\begin{tabular}{c p{32mm} c}
\toprule
\textbf{Function Index} & \textbf{Expression} & \textbf{Domain} \\
\midrule
1  & $f(x) = x + \ln(x^2 + 1)$ & \( x \in \mathbb{R} \) \\
2  & $f(x) = e^x + x^2$ & \( x \in \mathbb{R} \) \\
3  & $f(x) = x + \arctan(x)$ & \( x \in \mathbb{R} \) \\
4  & $f(x) = x\sqrt{x^2 + 1}$ & \( x \in \mathbb{R} \) \\
5  & $f(x) = x^3 + 3x$ & \( x \in \mathbb{R} \) \\
6  & $f(x) = x + \sqrt{x + 2} + \ln(x^2 + 1)$ & $x \geq -2$ \\
7  & $f(x) = e^x + \ln(x^2 + 1)$ & \( x \in \mathbb{R} \) \\
8  & $f(x) = \sqrt{x^2 + 1} + x^3$ & \( x \in \mathbb{R} \) \\
9  & $f(x) = x + \arcsin(\tanh(x))$ & \( x \in \mathbb{R} \) \\
10 & $f(x) = \ln(x + 2) + e^x$ & $x > -2$ \\
\bottomrule
\end{tabular}
\caption{ List of Continuous Globally Monotonically Increasing Functions }
\end{table}

\begin{table}[h!]
\centering
\renewcommand{\arraystretch}{1.2}  
\setlength{\tabcolsep}{12pt}  
\begin{tabular}{c p{32mm} c}
\toprule
\textbf{Function Index} & \textbf{Expressions} & \textbf{Domain} \\
\midrule
1  & \(f(x) = x \sin(x) \) & \( x \in \mathbb{R} \) \\
2  & \( f(x) = 3x^3 - 2x \) & \( x \in \mathbb{R} \) \\
3  & \( f(x) = \log(1+x) - \log(1-x) \) & \( x \in (-1, 1) \) \\
4  & \( f(x) = e^x - e^{-x} \) & \( x \in \mathbb{R} \) \\
5  & \( f(x) = \arctan(x) - \arctan(-x) \) & \( x \in \mathbb{R} \) \\
6  & \( f(x) = x(x^2 + 3) \) & \( x \in \mathbb{R} \) \\
7  & \( f(x) = x^5 - 10x^3 + 9x \) & \( x \in \mathbb{R} \) \\
8  & \( f(x) = \sinh(x) = \frac{e^x - e^{-x}}{2} \) & \( x \in \mathbb{R} \) \\
9  & \( f(x) = 7x - x^7 \) & \( x \in \mathbb{R} \) \\
10 & \( f(x) = x \cos(x) + \sin(x) \) & \( x \in \mathbb{R} \) \\
\bottomrule
\end{tabular}
\caption{List of Origin-Centered Symmetric (Odd Functions)}
\end{table}

\begin{table}[h!]
\centering
\renewcommand{\arraystretch}{1.2}  
\setlength{\tabcolsep}{12pt}  
\begin{tabular}{c p{32mm} c}
\toprule
\textbf{Function Index} & \textbf{Expression} & \textbf{Domain} \\
\midrule
1 & \( f(x) = x^4 + 2x^2 + 1 \) & \( x \in \mathbb{R} \) \\
2 & \( f(x) = e^x + x^2 \) & \( x \in \mathbb{R} \) \\
3 & \( f(x) = x^2 + \log(1 + e^x) \) & \( x \in \mathbb{R} \)\\
4 & \( f(x) = x \sinh(x) + \cosh(x) \) & \( x \in \mathbb{R} \) \\
5 & \( f(x) = x^2 + \sqrt{x^2 + 1} \) & \( x \in \mathbb{R} \) \\
6 & \( f(x) = e^{x^2} - x \) & \( x \in \mathbb{R} \) \\
7 & \( f(x) = x^2 + \arctan(x) \) & \( x \in \mathbb{R} \) \\
8 & \( f(x) = \sqrt{1 + e^x} \) & \( x \in \mathbb{R} \) \\
9 & \( f(x) = x + \log(1 + x^2) \) & \( x \in \mathbb{R} \)\\
10 & \( f(x) = x^6 + x^4 - x^3 + x + e^{-x} \) & \( x \in \mathbb{R} \)\\
\bottomrule
\end{tabular}
\caption{List of Complex Continuous Convex Functions}
\end{table}

\newpage

\section{Appendix: Test data in detail}
\label{AJ}
Tables \ref{a-tab1},\ref{a-tab2}, \ref{a-tab3}, and \ref{a-tab4} show in detail the expression forms of the data set used in the experiment.

\begin{table}[htbp]
\centering

\begin{scriptsize}
\begin{tabularx}{\linewidth}{cXc}
\toprule[1.45pt]
\toprule
Name & Expression & Dataset  \\ \hline
Korns-1 & $1.57+24.3*x_1^4$ & U$(-1, 1, 20)$  \\
Korns-2 & $0.23+14.2\frac{(x_4+x_1)}{(3x_2)}$ & U$(-1, 1, 20)$  \\
Korns-3 & $4.9\frac{(x_2-x_1+\frac{x_1}{x_3}}{(3x_3))}-5.41$ & U$(-1, 1, 20)$ \\
Korns-4 & $0.13sin(x_1)-2.3$ & U$(-1, 1, 20)$  \\
Korns-5 & $3+2.13log(|x_5|)$ & U$(-1, 1, 20)$  \\
Korns-6 & $1.3+0.13\sqrt{|x_1|}$ & U$(-1, 1, 20)$  \\
Korns-7 & $2.1(1-e^{-0.55x_1})$ & U$(-1,1 , 20)$  \\
Korns-8 & $6.87+11\sqrt{|7.23 x_1 x_4 x_5|}$ & U$(-1, 1, 20)$ \\
Korns-9 & $12\sqrt{|4.2x_1x_2x_2|}$ & U$(-1, 1, 20)$ \\
Korns-10 & $0.81+24.3\frac{2x_{1}+3x_2^2}{4x_3^3+5x_4^4}$ & U$(-1, 1, 20)$  \\
Korns-11 & $6.87+11cos(7.23x_1^3)$ & U$(-1, 1, 20)$  \\
Korns-12 & $2-2.1cos(9.8x_1^3)sin(1.3x_5)$ & U$(-1, 1, 20)$  \\ 
Korns-13 & $32.0-3.0\frac{tan(x_1)}{tan(x_2)}\frac{tan(x_3)}{tan(x_4)}$ & U$(-1, 1, 20)$ \\
Korns-14 & $22.0-(4.2cos(x_1)-tan(x_2))\frac{tanh(x_3)}{sin(x_4)}$ & U$(-1, 1, 20)$  \\
Korns-15 & $12.0-\frac{6.0tan(x_1)}{e^{x_2}}(log(x_3)-tan(x_4))))$ & U$(-1, 1, 20)$  \\ 
\toprule
\end{tabularx}
\end{scriptsize}
\caption{ Specific formula form and value range of the three data sets Korns. 
}
\label{a-tab1}
\end{table}

\begin{table}[htpb]
\centering
\begin{scriptsize}
\begin{tabularx}{\linewidth}{cXc}
\toprule[1.45pt]
\toprule
Name & Expression & Dataset \\
\hline
Neat-1 & $x_1^4+x_1^3+x_1^2+x$ & U$(-1, 1, 20)$  \\
Neat-2 & $x_1^5+x_1^4+x_1^3+x_1^2+x$ & U$(-1, 1, 20)$ \\
Neat-3 & $\sin(x_1^2)\cos(x)-1$ & U$(-1, 1, 20)$ \\
Neat-4 & $\log(x+1)+\log(x_1^2+1)$ & U$(0, 2, 20)$  \\
Neat-5 & $2\sin(x)\cos(x_2)$ & U$(-1, 1, 100)$  \\
Neat-6 & $\sum_{k=1}^x \frac{1}{k} $ & E$(1, 50, 50)$  \\
Neat-7 & $2 - 2.1\cos(9.8x_1)\sin(1.3x_2)$ & E$(-50, 50, 10^5)$ \\
Neat-8 & $\frac{e^{-(x_1)^2}}{1.2 + (x_2-2.5)^2}$ & U$(0.3, 4, 100)$  \\
Neat-9 & $\frac{1}{1+x_1^{-4}} + \frac{1}{1+x_2^{-4}}$ & E$(-5, 5, 21)$ \\
\toprule
Keijzer-1 & $0.3x_1sin(2\pi x_1)$ & U$(-1, 1, 20)$  \\
Keijzer-2 & $2.0x_1sin(0.5\pi x_1)$ & U$(-1, 1, 20)$  \\
Keijzer-3 & $0.92x_1sin(2.41\pi x_1)$ & U$(-1, 1, 20)$ \\
Keijzer-4 & $x_1^3e^{-x_1}cos(x_1)sin(x_1)sin(x_1)^{2}cos(x_1)-1$ & U$(-1, 1, 20)$ \\
Keijzer-5 & $3+2.13log(|x_5|)$ & U$(-1, 1, 20)$\\

Keijzer-6 & $\frac{x1(x1+1)}{2}$& U$(-1, 1, 20)$ \\
Keijzer-7 & $log(x_1)$ & U$(0,1 , 20)$ \\
Keijzer-8 & $\sqrt{(x_1)}$ & U$(0, 1, 20)$  \\
Keijzer-9 & $log(x_1+\sqrt{x_1^2}+1)$ & U$(-1, 1, 20)$ \\
Keijzer-10 & $x_{1}^{x_2}$ & U$(-1, 1, 20)$  \\
Keijzer-11 & $x_1x_2+sin((x_1-1)(x_2-1))$ & U$(-1, 1, 20)$  \\
Keijzer-12 & $x_1^4-x_1^3+\frac{x_2^2}{2}-x_2$ & U$(-1, 1, 20)$  \\ 
Keijzer-13 & $6sin(x_1)cos(x_2)$ & U$(-1, 1, 20)$  \\
Keijzer-14 & $\frac{8}{2+x_1^2 + x_2^2}$ & U$(-1, 1, 20)$ \\
Keijzer-15 & $\frac{x_1^3}{5}+\frac{x_2^3}{2}-x_2-x_1$ & U$(-1, 1, 20)$ \\ 

\toprule
Livermore-1 & $\frac{1}{3}+x_1+sin(x_1^2))$ & U$(-3, 3, 100)$  \\
Livermore-2 & $sin(x_1^2)*cos(x1)-2$ & U$(-3, 3, 100)$  \\
Livermore-3 & $sin(x_1^3)*cos(x_1^2))-1$ & U$(-3, 3, 100)$  \\
Livermore-4 & $log(x_1+1)+log(x_1^2+1)+log(x_1)$ & U$(-3, 3, 100)$ \\ 
Livermore-5 & $x_1^4-x_1^3+x_2^2-x_2$ & U$(-3, 3, 100)$  \\
Livermore-6 & $4x_1^4+3x_1^3+2x_1^2+x_1$ & U$(-3, 3, 100)$ \\ 
Livermore-7 & $\frac{(exp(x1)-exp(-x_1)}{2})$ & U$(-1, 1, 100)$ \\ 
Livermore-8 & $\frac{(exp(x1)+exp(-x1)}{3}$ & U$(-3, 3, 100)$ \\
Livermore-9 & $x_1^9+x_1^8+x_1^7+x_1^6+x_1^5+x_1^4+x_1^3+x_1^2+x_1$ & U$(-1, 1, 100)$  \\
Livermore-10 & $6*sin(x_1)cos(x_2)$ & U$(-3, 3, 100)$  \\
Livermore-11 & $\frac{x_1^2 x_2^2}{(x_1+x_2)}$ & U$(-3, 3, 100)$ \\
Livermore-12 & $\frac{x_1^5}{x_2^3}$ & U$(-3, 3, 100)$  \\
Livermore-13 & $x_1^{\frac{1}{3}}$ & U$(-3, 3, 100)$  \\
Livermore-14 & $x_1^3+x_1^2+x_1+sin(x_1)+sin(x_2^2)$ & U$(-1, 1, 100)$ \\ 
Livermore-15 & $x_1^\frac{1}{5}$ & U$(-3, 3, 100)$  \\
Livermore-16 & $x_1^{\frac{2}{3}}$ & U$(-3, 3, 100)$  \\  
Livermore-17 & $4sin(x_1)cos(x_2)$ & U$(-3, 3, 100)$  \\
Livermore-18 & $sin(x_1^2)*cos(x_1)-5$ & U$(-3, 3, 100)$  \\
Livermore-19 & $x_1^5+x_1^4+x_1^2 + x_1$ & U$(-3, 3, 100)$  \\
Livermore-20 & $e^{(-x_1^2)}$ & U$(-3, 3, 100)$  \\
Livermore-21 & $x_1^8+x_1^7+x_1^6+x_1^5+x_1^4+x_1^3+x_1^2+x_1$& U$(-1, 1, 20)$ \\
Livermore-22 & $e^{(-0.5x_1^2)}$ & U$(-3, 3, 100)$  \\
\toprule
\end{tabularx}
\end{scriptsize}
\caption{
Specific formula form and value range of the three data sets neat, Keijzer, and Livermore.
}
\label{a-tab2}
\end{table}

\begin{table}[htbp]
\centering

\begin{scriptsize}
\begin{tabularx}{\linewidth}{cXc}
\toprule[1.45pt]
\toprule
Name & Expression & Dataset  \\ \hline
Nguyen-1 & $x_1^3+x_1^2+x_1$&U$(-1, 1, 20)$\\
Nguyen-2 & $x_1^4+x_1^3+x_1^2+x_1$ & U$(-1, 1, 20)$ \\
Nguyen-3 & $x_1^5+x_1^4+x_1^3+x_1^2+x_1$ & U$(-1, 1, 20)$ \\
Nguyen-4 & $x_1^6+x_1^5+x_1^4+x_1^3+x_1^2+x_1$ & U$(-1, 1, 20)$  \\
Nguyen-5 & $\sin(x_1^2)\cos(x)-1$ & U$(-1, 1, 20)$  \\
Nguyen-6 & $\sin(x_1)+\sin(x_1+x_1^2)$ & U$(-1, 1, 20)$  \\
Nguyen-7 & $\log(x_1+1)+\log(x_1^2+1)$ & U$(0, 2, 20)$  \\
Nguyen-8 & $\sqrt{x}$ & U$(0, 4, 20)$  \\
Nguyen-9 & $\sin(x)+\sin(x_2^2)$ & U$(0, 1, 20)$ \\
Nguyen-10 & $2\sin(x)\cos(x_2)$ & U$(0, 1, 20)$ \\
Nguyen-11 & $x_1^{x_2}$ & U$(0, 1, 20)$  \\
Nguyen-12 & $x_1^4-x_1^3+\frac{1}{2}x_2^2-x_2$ & U$(0, 1, 20)$ \\

\toprule
Nguyen-2$'$ & $4x_1^4+3x_1^3+2x_1^2+x$ & U$(-1, 1, 20)$  \\
Nguyen-5$'$ & $\sin(x_1^2)\cos(x)-2$ & U$(-1, 1, 20)$  \\
Nguyen-8$'$ & $\sqrt[3]{x}$ & U$(0, 4, 20)$ \\
Nguyen-8$''$ & $\sqrt[3]{x_1^2}$ & U$(0, 4, 20)$ \\
\toprule
Nguyen-1\textsuperscript{c} & $3.39x_1^3+2.12x_1^2+1.78x$ & U$(-1, 1, 20)$ \\
Nguyen-5\textsuperscript{c} & $\sin(x_1^2)\cos(x)-0.75$ & $U(-1, 1, 20)$  \\
Nguyen-7\textsuperscript{c} & $\log(x+1.4)+\log(x_1^2+1.3)$ & U$(0, 2, 20)$ \\
Nguyen-8\textsuperscript{c} & $\sqrt{1.23 x}$ & U$(0, 4, 20)$  \\
Nguyen-10\textsuperscript{c} & $\sin(1.5x)\cos(0.5x_2)$ & U$(0, 1, 20)$  \\

\toprule
Jin-1 & $2.5 x_1^4-1.3 x_1^3 +0.5 x_2^2 - 1.7x_2$ & U$(-3, 3, 100)$ \\
Jin-2 & $8.0 x_1^2 + 8.0 x_2^3 - 15.0$ & U$(-3, 3, 100)$  \\
Jin-3 & $0.2 x_{1}^{3} + 0.5 x_{2}^{3} - 1.2 x_2 - 0.5 x_{1}$ & U$(-3, 3, 100)$  \\    
Jin-4 & $1.5 \exp{x} + 5.0 cos(x_2)$ & U$(-3, 3, 100)$\\
Jin-5 & $6.0 sin(x_1) cos(x_2)$ & U$(-3, 3, 100)$ \\
Jin-6 & $1.35 x_1 x_2 + 5.5 sin((x_1 - 1.0)(x_2 - 1.0))$ & U$(-3, 3, 100)$ \\   
\toprule
\end{tabularx}
\end{scriptsize}
\caption{ Specific formula form and value range of the three data sets Nguyen, and Jin. 
}
\label{a-tab3}
\end{table}

\begin{table}[htpb]
\centering
\begin{scriptsize}
\begin{tabularx}{\linewidth}{cXc}
\toprule[1.45pt]
\toprule
Name & Expression & Dataset \\
\toprule
Vladislavleva-1 & $\frac{(e^{-(x1-1)^2})}{(1.2+(x2-2.5)^2))}$ & U$(-1, 1, 20)$ \\
Vladislavleva-2 & $e^{-x_1}x_1^3cos(x_1)sin(x_1)(cos(x_1)sin(x_1)^2-1)$ & U$(-1, 1, 20)$ \\

Vladislavleva-3 & $e^{-x_1}x_1^3cos(x_1)sin(x_1)(cos(x_1)sin(x_1)^2-1)(x_2-5)$ & U$(-1, 1, 20)$ \\
Vladislavleva-4 & $\frac{10}{5+(x1-3)^2+(x_2-3)^2+(x_3-3)^2+(x_4-3)^2+(x_5-3)^2}$ & U$(0, 2, 20)$ \\
Vladislavleva-5 & $30(x_1-1)\frac{x_3-1}{(x_1-10)}x_2^2$ & U$(-1, 1, 100)$ \\
Vladislavleva-6 & $6sin(x_1)cos(x_2)$ & E$(1, 50, 50)$ \\
Vladislavleva-7 & $2 - 2.1\cos(9.8x)\sin(1.3x_2)$ & E$(-50, 50, 10^5)$ \\
Vladislavleva-8 & $\frac{e^{-(x-1)^2}}{1.2 + (x_2-2.5)^2}$ & U$(0.3, 4, 100)$  \\
\toprule
Test-2 & $3.14x_1^2$ & U$(-1, 1, 20)$ \\
Const-Test-1 & $5x_1^2$ & U$(-1, 1, 20)$ \\
GrammarVAE-1 & $1/3+x1+sin(x_1^2))$ & U$(-1, 1, 20)$ \\
Sine & $sin(x_1)+sin(x_1+x_1^2))$ & U$(-1, 1, 20)$ \\
Nonic & $x_1^9+x_1^8+x_1^7+x_1^6+x_1^5+x_1^4+x_1^3+x_1^2+x_1$ & U$(-1, 1, 100)$  \\
Pagie-1 & $\frac{1}{1+x_1^{-4}+\frac{1}{1+x2^{-4}}} $ & E$(1, 50, 50)$  \\
Meier-3 & $\frac{x_1^2  x_2^2}{(x_1+x_2)}$ & E$(-50, 50, 10^5)$ \\
Meier-4 & $\frac{x_1^5}{x_2^3}$ & $U(0.3, 4, 100)$  \\
Poly-10 & $x_1x_2+x_3x4+x_5x_6+x_1x_7x_9+x_3x_6x_{10}$ & E$(-1, 1, 100)$ \\
\toprule
Constant-1 & $3.39*x_1^3+2.12*x_1^2+1.78*x_1$&$U(-4, 4, 100)$\\
Constant-2 & $sin(x_1^2)*cos(x_1)-0.75$&$U(-4, 4, 100)$\\
Constant-3 & $sin(1.5*x_1)*cos(0.5*x_2)$&$U(0.1, 4, 100)$\\
Constant-4 & $2.7*x_1^{x_2}$&$U(0.3, 4, 100)$\\
Constant-5 & $sqrt(1.23*x_1)$&$U(0.1, 4, 100)$\\
Constant-6 & $x_1^{0.426}$&$U(0.0, 4, 100)$\\
Constant-7 & $2*sin(1.3*x_1)*cos(x_2)$&$U(-4, 4, 100)$\\
Constant-8 & $log(x_1+1.4)+log(x1,2+1.3)$&$U(-4, 4, 100)$\\
\toprule
R1 & $\frac{(x_1+1)^3}{x_1^2-x_1+1)}$&$U(-5, 5, 100)$\\
R2 & $\frac{(x_1^2-3*x_1^2+1}{x_1^2+1)}$&$U(-4, 4, 100)$\\
R3 & $\frac{x_1^6+x_1^5)}{(x_1^4+x_1^3+x_1^2+x1+1)}$&$U(-4, 4, 100)$\\
\toprule
\end{tabularx}
\end{scriptsize}
\caption{
Specific formula form and value range of the three data sets, Vladislavleva and others. }
\label{a-tab4}
\end{table}
\section{Appendix: ChatSR tests on AIFeynman dataset.}
\label{AK}
In our study, we conducted an evaluation of our novel symbol regression algorithm, termed ChatSR, leveraging the AI Feynman dataset. Detailed experimental results are presented in Table \ref{a-tab5} and Table \ref{a-tab6}.
\begin{table}[htbp]
\centering
{\footnotesize
\begin{tabular}{|l|l|r|}
\hline
Feynman   & Equation & $R^2$ \\
\hline                            
I.6.20a       & $f = e^{-\theta^2/2}/\sqrt{2\pi}$ & 0.9999  \\
I.6.20        & $f = e^{-\frac{\theta^2}{2\sigma^2}}/\sqrt{2\pi\sigma^2}$ & 0.9983\\
I.6.20b       & $f = e^{-\frac{(\theta-\theta_1)^2}{2\sigma^2}}/\sqrt{2\pi\sigma^2}$ & 0.9934 \\
I.8.14       & $d = \sqrt{(x_2-x_1)^2+(y_2-y_1)^2}$ & 0.9413  \\
I.9.18       & $F = \frac{Gm_1m_2}{(x_2-x_1)^2+(y_2-y_1)^2+(z_2-z_1)^2}$  & 0.9938\\
I.10.7       & $F = \frac{Gm_1m_2}{(x_2-x_1)^2+(y_2-y_1)^2+(z_2-z_1)^2}$  & 0.9782\\
I.11.19      & $A = x_1y_1+x_2y_2+x_3y_3$ & 0.9993   \\
I.12.1       & $F = \mu N_n$ & 0.9999 \\
I.12.2       & $F = \frac{q_1q_2}{4\pi\epsilon r^2}$   & 0.9999 \\
I.12.4       & $E_f = \frac{q_1}{4\pi\epsilon r^2}$  & 0.9999 \\
I.12.5       & $F = q_2 E_f$ & 0.9999  \\
I.12.11      & $F = \mathcal{Q}(E_f+B v \sin\theta)$  & 0.9969 \\
I.13.4      & $K = \frac{1}{2}m(v^2+u^2+w^2)$  & 0.9831  \\
I.13.12      & $U = Gm_1m_2(\frac{1}{r_2}-\frac{1}{r_1})$ & 0.9999  \\
I.14.3       & $U = mgz$ &1.0    \\
I.14.4       & $U = \frac{k_{spring}x^2}{2}$  & 0.9924  \\
I.15.3x      & $x_1 = \frac{x-ut}{\sqrt{1-u^2/c^2}}$ & 0.9815 \\
I.15.3t      & $t_1 = \frac{t-ux/c^2}{\sqrt{1-u^2/c^2}}$ & 0.9822  \\
I.15.10       & $p = \frac{m_0v}{\sqrt{1-v^2/c^2}}$ & 0.9920 \\
I.16.6       & $v_1 = \frac{u+v}{1+uv/c^2}$ & 0.9903  \\
I.18.4       & $r = \frac{m_1r_1+m_2r_2}{m_1+m_2}$ & 0.9711 \\
I.18.12      & $\tau = rF\sin\theta$  & 0.9999  \\
I.18.16      & $L = mrv \sin\theta$  & 0.9997 \\
I.24.6 & $E = \frac{1}{4} m (\omega^2+\omega_0^2) x^2$      & 0.9991\\
I.25.13      & $V_e = \frac{q}{C}$ & 1.0 \\
I.26.2       & $\theta_1 = \arcsin(n  \sin\theta_2)$ & 0.9989 \\
I.27.6       & $f_f$    $ = \frac{1}{\frac{1}{d_1}+\frac{n}{d_2}}$  & 0.9942 \\
I.29.4       & $k = \frac{\omega}{c}$ & 1.0 \\
I.29.16      & $x = \sqrt{x_1^2+x_2^2-2x_1x_2\cos(\theta_1-\theta_2)}$ & 0.9922  \\
I.30.3 & $I_* = I_{*_0}\frac{\sin^2(n\theta/2)}{\sin^2(\theta/2)}$ & 0.9946 \\
I.30.5       & $\theta = \arcsin(\frac{\lambda}{nd})$  & 0.9933\\
I.32.5       & $P = \frac{q^2a^2}{6\pi\epsilon c^3}$       & 0.9905 \\
I.32.17 & $P = (\frac{1}{2}\epsilon c E_f^2)(8\pi r^2/3) (\omega^4/(\omega^2-\omega_0^2)^2)$      & 0.9941  \\
I.34.8       & $\omega = \frac{qvB}{p}$   & 1.0\\
I.34.10       & $\omega = \frac{\omega_0}{1-v/c}$ & 0.9903 \\
I.34.14      & $\omega = \frac{1+v/c}{\sqrt{1-v^2/c^2}}\omega_0$  & 0.9941 \\
I.34.27      & $E = \hbar\omega$  & 0.9999 \\
I.37.4       & $I_* = I_1+I_2+2\sqrt{I_1I_2}\cos\delta$ & 0.9723\\
I.38.12      & $r = \frac{4\pi\epsilon\hbar^2}{mq^2}$   & 0.9999  \\
I.39.10       & $E = \frac{3}{2}p_F V$     & 0.9981 \\
I.39.11      & $E = \frac{1}{\gamma-1}p_F V$  & 0.9883 \\
I.39.22      & $P_F = \frac{n k_b T}{V}$       & 0.9902  \\
I.40.1       & $n = n_0e^{-\frac{mgx}{k_bT}}$    & 0.9924 \\
I.41.16      & $L_{rad} = \frac{\hbar\omega^3}{\pi^2c^2(e^{\frac{\hbar\omega}{k_bT}}-1)}$ & 0.9435  \\
I.43.16      & $v = \frac{\mu_{drift}q V_e}{d}$   & 0.9903  \\
I.43.31      & $D = \mu_e k_bT$    & 1.0  \\
I.43.43      & $\kappa = \frac{1}{\gamma-1}\frac{k_bv}
{A}$  & 0.9333  \\
I.44.4       & $E = n k_b T \ln(\frac{V_2}{V_1})$   & 0.8624  \\
I.47.23      & $c = \sqrt{\frac{\gamma pr}{\rho}}$   & 0.9624 \\
I.48.20       & $E = \frac{m c^2}{\sqrt{1-v^2/c^2}}$ &  0.8866\\
I.50.26 & $x = x_1[\cos(\omega t)+\alpha\> cos(\omega t)^2]$      & 0.9999   \\
\hline
\end{tabular}
\caption{Tested Feynman Equations, part 1.}
\label{a-tab5}
}
\end{table}
\begin{table}[htbp]
\centering
\renewcommand{\arraystretch}{1.0} 
{\footnotesize

\begin{tabular}{|l|l|r|}
\hline
Feynman   & Equation & $R^2$\\
\hline       
II.2.42   & P     $ = \frac{\kappa(T_2-T_1)A}{d}$  & 0.8335  \\
II.3.24   & $F_E = \frac{P}{4\pi r^2}$  & 0.9755 \\
II.4.23   & $V_e = \frac{q}{4\pi\epsilon r}$   & 0.9901 \\
II.6.11 & $V_e =\frac{1}{4\pi\epsilon}\frac{p_d\cos \theta}{r^2}$      & 0.9913 \\
II.6.15a & $E_f = \frac{3}{4\pi\epsilon}\frac{p_d z}{r^5} \sqrt{x^2+y^2}$      & 0.9031  \\
II.6.15b & $E_f = \frac{3}{4\pi\epsilon}\frac{p_d}{r^3} \cos\theta\sin\theta$      & 0.9925  \\
II.8.7    & $E = \frac{3}{5}\frac{q^2}{4\pi\epsilon d}$  & 0.9736  \\
II.8.31   & $E_{den} = \frac{\epsilon E_f^2}{2}$                     & 0.9999 \\
II.10.9   & $E_f = \frac{\sigma_{den}}{\epsilon}\frac{1}{1+\chi}$      & 0.9939  \\
II.11.3 & $x = \frac{q E_f}{m(\omega_0^2-\omega^2)}$      & 0.9903     \\
II.11.7 & $n = n_0(1+ \frac{p_d E_f \cos\theta}{k_b T})$      & 0.8305 \\
II.11.20  & $P_* = \frac{n_\rho p_d^2 E_f}{3 k_b T}$ & 0.8013  \\
II.11.27 & $P_* = \frac{n\alpha}{1-n\alpha/3}\epsilon E_f$      & 0.9816   \\
II.11.28  & $\theta = 1+\frac{n\alpha}{1-(n\alpha/3)}$    & 0.9985\\ 
II.13.17  & $B = \frac{1}{4 \pi \epsilon c^2}\frac{2I}{r}$ & 0.9991\\
II.13.23  & $\rho_c = \frac{\rho_{c_0}}{\sqrt{1-v^2/c^2}}$          & 0.9832  \\
II.13.34  & $j = \frac{\rho_{c_0}v}{\sqrt{1-v^2/c^2}}$     & 0.9747 \\
II.15.4   & $E = -\mu_M B \cos\theta$               & 0.9999 \\
II.15.5   & $E = -p_d E_f\cos\theta$  & 0.9964 \\
II.21.32  & $V_e = \frac{q}{4\pi\epsilon r(1-v/c)}$   & 0.9899   \\
II.24.17 & $k = \sqrt{\frac{\omega^2}{c^2}-\frac{\pi^2}{d^2}}$      & 0.9835   \\
II.27.16  & $F_E = \epsilon c E_f^2$        & 0.9954 \\
II.27.18  & $E_{den} = \epsilon E_f^2$         & 0.9952 \\
II.34.2a  & $I = \frac{qv}{2\pi r}$         & 0.9835 \\
II.34.2   & $\mu_M = \frac{q v r}{2}$             & 0.9946 \\
II.34.11  & $\omega = \frac{g_{\_} q B}{2m}$          & 0.9935 \\
II.34.29a & $\mu_M = \frac{q h}{4\pi m}$      & 0.9956  \\
II.34.29b & $E = \frac{g_{\_} \mu_M B J_z}{\hbar}$ & 0.8614\\
II.35.18 & $n = \frac{n_0}{\exp(\mu_m B/(k_b T))+\exp(-\mu_m B/(k_b T))}$      & 0.9647 \\
II.35.21  & $M = n_\rho \mu_M \tanh(\frac{\mu_M B}{k_b T})$     & 0.8097 \\
II.36.38 & $f = \frac{\mu_m B}{k_b T}+\frac{\mu_m\alpha M}{\epsilon c^2 k_b T}$      & 0.9840\\
II.37.1   & $E = \mu_M(1+\chi)B$    & 0.9999\\
II.38.3   & $F = \frac{Y A x}{d}$            & 0.9979 \\
II.38.14  & $\mu_S = \frac{Y}{2(1+\sigma)}$     & 0.9999  \\
III.4.32  & $n = \frac{1}{e^{\frac{\hbar\omega}{k_bT}}-1}$ & 0.9812  \\
III.4.33  & $E = \frac{\hbar\omega}{e^{\frac{\hbar\omega}{k_b T}}-1}$  & 0.9964    \\
III.7.38  & $\omega = \frac{2 \mu_M B}{\hbar}$  & 0.9932  \\
III.8.54  & $p_{\gamma}$    $ = \sin(\frac{E t}{\hbar})^2$  & 0.9943\\
III.9.52  & $p_{\gamma}$    $ = \frac{p_d E_f t}{\hbar} \frac{    \sin((\omega-\omega_0)t/2)^2}{((\omega-\omega_0)t/2)^2}$ & 0.7622  \\
III.10.19 & $E = \mu_M\sqrt{B_x^2+B_y^2+B_z^2}$  & 0.9964 \\
III.12.43 & $L = n\hbar$ & 0.9993  \\
III.13.18 & $v = \frac{2 E d^2 k}{\hbar}$ & 0.9959  \\
III.14.14 & $I = I_0 (e^{\frac{q V_e}{k_b T}}-1)$  & 0.9925\\
III.15.12 & $E = 2U(1-\cos(kd))$    & 0.9998 \\
III.15.14 & $m = \frac{\hbar^2}{2E d^2}$     & 0.9947  \\
III.15.27 & $k = \frac{2\pi\alpha}{nd}$    & 0.9992 \\
III.17.37 & $f = \beta(1+\alpha \cos\theta)$ & 0.9925 \\
III.19.51 & $E = \frac{-mq^4}{2(4\pi\epsilon)^2\hbar^2}\frac{1}{n^2}$     & 0.9934 \\
III.21.20 & $j = \frac{-\rho_{c_0} q A_{vec}}{m}$  & 0.8036  \\
\hline
\end{tabular} 
\caption{Tested Feynman Equations, part 2.}
\label{a-tab6}
}
\end{table}

\end{document}